\documentclass[sigconf,nonacm]{aamas}

\usepackage{balance} 

\usepackage{graphicx}

\usepackage{subfigure}
\usepackage{amsmath}
\usepackage{mathtools}
\usepackage{amsthm}
\usepackage{soul}
\usepackage{float}

\usepackage{wrapfig}
\usepackage{algpseudocode}
\usepackage{algorithm}
\usepackage{setspace}

\newtheorem{theorem}{Theorem}

\newtheorem{remark}[theorem]{Remark}

\copyrightyear{}
\acmYear{}
\acmDOI{}
\acmPrice{}
\acmISBN{}

\acmSubmissionID{790}

\title[AAMAS-2024 Formatting Instructions]{Fully Independent Communication in Multi-Agent Reinforcement Learning}

\author{Rafael Pina}
\affiliation{
  \institution{Loughborough University London}
  \city{London}
  \country{United Kingdom}}
\email{r.m.pina@lboro.ac.uk}

\author{Varuna De Silva}
\affiliation{
  \institution{Loughborough University London}
  \city{London}
  \country{United Kingdom}}
\email{v.d.de-silva@lboro.ac.uk}

\author{Corentin Artaud}
\affiliation{
  \institution{Loughborough University London}
  \city{London}
  \country{United Kingdom}}
\email{c.artaud2@lboro.ac.uk}

\author{Xiaolan Liu}
\affiliation{
  \institution{Loughborough University London}
  \city{London}
  \country{United Kingdom}}
\email{xiaolan.liu@lboro.ac.uk}

\begin{abstract}
Multi-Agent Reinforcement Learning (MARL) comprises a broad area of research within the field of multi-agent systems. Several recent works have focused specifically on the study of communication approaches in MARL. While multiple communication methods have been proposed, these might still be too complex and not easily transferable to more practical contexts. One of the reasons for that is due to the use of the famous parameter sharing trick. In this paper, we investigate how independent learners in MARL that do not share parameters can communicate. We demonstrate that this setting might incur into some problems, to which we propose a new learning scheme as a solution. Our results show that, despite the challenges, independent agents can still learn communication strategies following our method. Additionally, we use this method to investigate how communication in MARL is affected by different network capacities, both for sharing and not sharing parameters. We observe that communication may not always be needed and that the chosen agent network sizes need to be considered when used together with communication in order to achieve efficient learning.
\end{abstract}


\newcommand{\BibTeX}{\rm B\kern-.05em{\sc i\kern-.025em b}\kern-.08em\TeX}

\begin{document}


\pagestyle{fancy}
\fancyhead{}


\maketitle 


\section{Introduction}

Communication in Multi-Agent Reinforcement Learning (MARL) has been an important topic of research in the broad field of MARL \cite{jiang_learning_2018,foerster_learning_2016,das_tarmac_2019,kim2019learning,kim_communication_2021}. Usually, in the standard approaches, agents learn the tasks by observing certain parts of the environment around them, and then make a decision based on what they see. However, if the agents are carriers of communication capabilities, they can use other information besides only their own observations to make a better decision. For instance, they can receive observations from the other agents, or their actions \cite{das_tarmac_2019,kim_communication_2021}.   

From the perspective of practical applications, communication is also seen as a feasible way of improving learning due to progresses in diverse fields \cite{furda_comm_driverless_2010,chen_wireless_apps_2019,mao2019learning}. For example, in scenarios of autonomous vehicles or factories with multiple agents, communication can be key to learn better responses to events that some agents can see at a certain moment, but the others cannot \cite{mason2023multi}. Having this prior knowledge of certain occurrences can help to prevent catastrophic events.  

In conventional MARL approaches, communication can often be applied by adding an additional network that learns how to produce messages that follow a certain communication protocol \cite{liu_multi-agent_2021,zhang2019efficient,ding2022sequential}. This network can be integrated into the learning process of the agents and improve the performance of the standard MARL algorithm. In simple terms, each agent produces a certain message that represents its own knowledge or experience at a certain moment, and then this message is broadcasted to the other agents. After it is delivered, the message is used as an additional input to the standard networks of the agents, meaning that now they have some knowledge about something else that was sent by the others, although it is encoded. 

In most MARL approaches, it is adopted a configuration where the parameters of the learning networks are shared \cite{rashid_qmix_2018,wang2021qplex,kim2019learning,gupta_2017,sunehag2017valuedecomposition}. This setting is often referred to simply as parameter sharing and, as the name suggests, approaches that adopt this configuration use only a single network (or two, if there is a mixer network or a communication network, for instance) that is shared by all the agents of the team. However, when we look at practical applications, sharing parameters becomes unrealistic \cite{wong2023deep}. Within the multiple proposed communication methods, when sharing parameters is not feasible a question naturally arises: can communication still be conducted successfully when the agents do not share parameters? In this paper, we investigate communication among independent learners in MARL who do not share the parameters of their networks and consider agents that have distinct networks for their policy and for generating communication messages. We demonstrate that this kind of communication can be challenging to achieve due to the parameters of the networks not being shared, but it still is possible thanks to a new learning scheme that we propose in this paper. Additionally, in the course of the experiments, we argue that communication in MARL might not always be necessary and, if it is used naively without considering the environment, it ends up bringing overhead to the learning process, without any benefits. 

To further analyse the effect of communication in the learning process of MARL, we also investigate whether simply increasing the capacity of the networks of the agents can compensate for the absence of a communication network. This is because, by increasing the capacity of the agent network, the amount of information that this network can represent also increases, which could accommodate for the absence of communication. On the other hand, communication enables the flow of information across agents, which, at first glance, should always be beneficial, posing another interesting question.    

Overall, in this paper, we intend to study the challenges of communication in independent MARL when parameters are not shared, which is an understudied setting in MARL that can bring benefits for practical applications \cite{wong2023deep}. In this sense, we propose a way of successfully communicating under these conditions. We also show that communication might not always be useful, bringing useless overhead, and show how it affects learning when the agent networks have higher or lower capacities.

\section{Background}
\label{sec:bkgd}
\subsection{Decentralised Partially Observable Markov Decision Processes (Dec-POMDPs)}
\sloppy In this work we formalise the treated scenarios following Decentralised Partially Observable Markov Decision Processes (Dec-POMDPs) \cite{oliohek_dec_pomdp_2016}. These can be represented by the tuple $G\mathrm{=}\left\langle S,A,r,O,Z,P,N\mathrm{,\ }\gamma \right\rangle$. At each state $s \in S$, each agent $i\in \mathcal{N}\mathrm{\equiv }\mathrm{\{}\mathrm{1,\dots ,}N\}$ chooses an action $a \in A$, forming a joint action $\equiv=\{a_1,\dots,a_N\}$ that is executed as a whole in the environment, resulting in a transition to a next state $s'$ according to a probability $P(s'|s,a):S\times A\times S\rightarrow\left[0,1\right]$. This results in a reward $r(s,a):S\times A\rightarrow\mathbb{R}$ that is shared by the team. Because of partial observability, at each state each agent receives only a local observation $o_i \in O(s,i):S \times \mathcal{N} \rightarrow Z$, and each agent holds an action-observation history $\tau_i \in \mathcal{T}\equiv(Z\times A)^*$. In the context of communication, each agent generates a certain message $m_i \in \mathcal{M}$ that is then broadcasted to the others and will also condition their policies. Thus, if a set of incoming messages to agent $i$ except their own is represented by $m_{-i}$, then its policy can be represented as $\pi_i(a_i|\tau_i,m_{-i})$. The joint policy aims to optimise a joint action-value function $Q_{\pi}(s_t,a_t)=\mathbb{E}_{\pi}[R_t|s_t,a_t]$, where $R_t = \sum_{k=0}^\infty \gamma^kr_{t+k}$ is the discounted return and $\gamma \in [0,1)$ is a discount factor.

\subsection{Independent Deep Q-Learning (IQL)}\label{sec:bkg_iql}
As one of the first proposed approaches for multi-agent learning, Independent Q-learning (IQL) can be seen as the most straightforward MARL method \cite{tan_multi-agent_1993}. In simpler terms, IQL consists of generalising the concepts from single-agent reinforcement learning to multi-agent settings, i.e., each agent produces an individual Q-function that is updated following the equation
\begin{equation}\label{eq:q_up}
    Q(s,a)=(1-\alpha)Q(s,a)+\alpha\left[r+\gamma\mathop{\mathrm{max}}_{a'}Q(s',a')\right],
\end{equation}
where $\alpha$ is a learning rate. Following the introduction of Deep Q-Networks (DQN) \cite{Mnih2015HumanlevelCT}, in the work of \cite{tampuu2015multiagent} the authors use together the advances from deep reinforcement learning and independent Q-learning to propose an improved independent Q-learning where the agents are now controlled by individual deep Q-networks instead of following only the simpler tabular case. Importantly, in DQN the authors introduce an experience replay buffer that is kept by the learning agent and the use of a target network that stabilises learning. Logically, this can be replicated in the multi-agent case, where each agent maintains all these components on its own. For brevity, in this paper we refer to this method also as Independent Q-learning (IQL), and use deep recurrent Q-networks, as introduced in \cite{hausknecht2017deep}, to accommodate for partial observability. Overall, IQL is trained to minimise the loss
\begin{equation}\label{eq:vff_loss}
    \mathcal{L}(\theta)=\mathbb{E}_{b\sim B}\left[\big(r+\gamma\mathop{\mathrm{max}}_{a'}Q(\tau',a';\theta^-)-Q(\tau,a;\theta)\big)^2\right],
\end{equation}
for a sample $b$ sampled from a replay buffer of experiences $B$, and where $\theta$ and $\theta^-$ are the parameters of the learning network and a target network, respectively.
\begin{figure}[!t]
    \centering 
    \subfigure[Parameter Sharing]{\label{fig:env_a}\includegraphics[width=0.48\columnwidth]{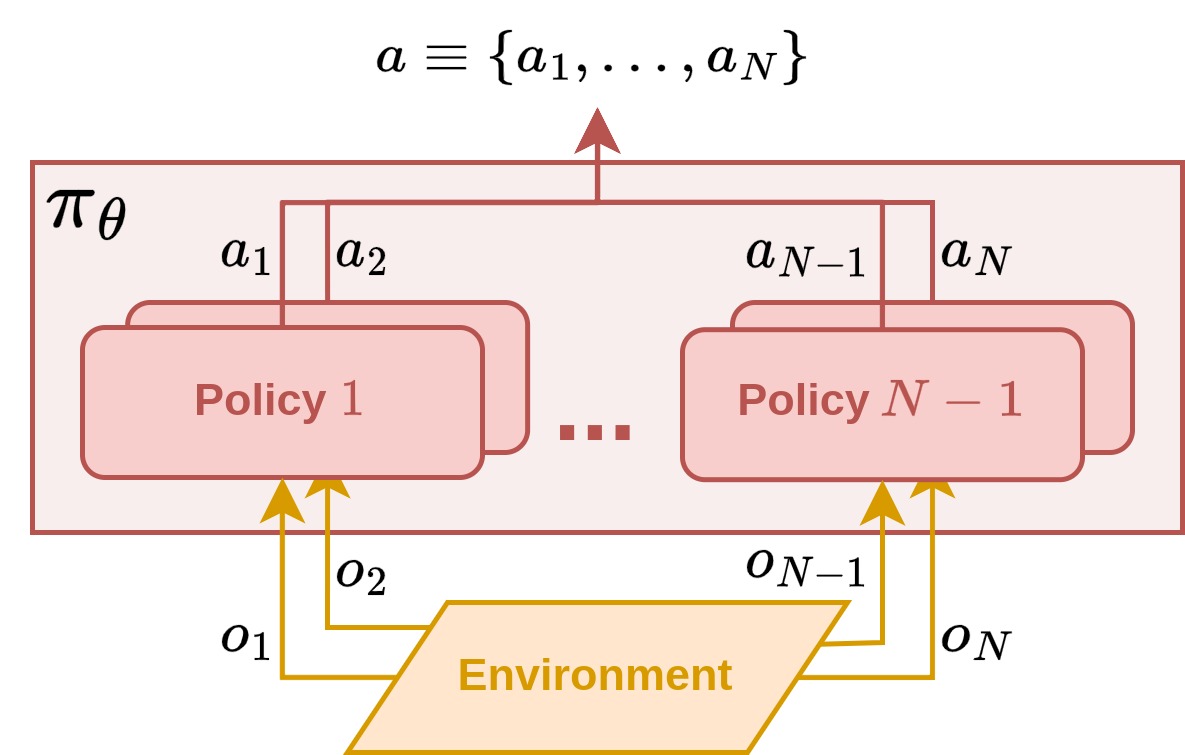}}
    \subfigure[\textbf{No} Parameter Sharing]{\label{fig:env_b}\includegraphics[width=0.48\columnwidth]{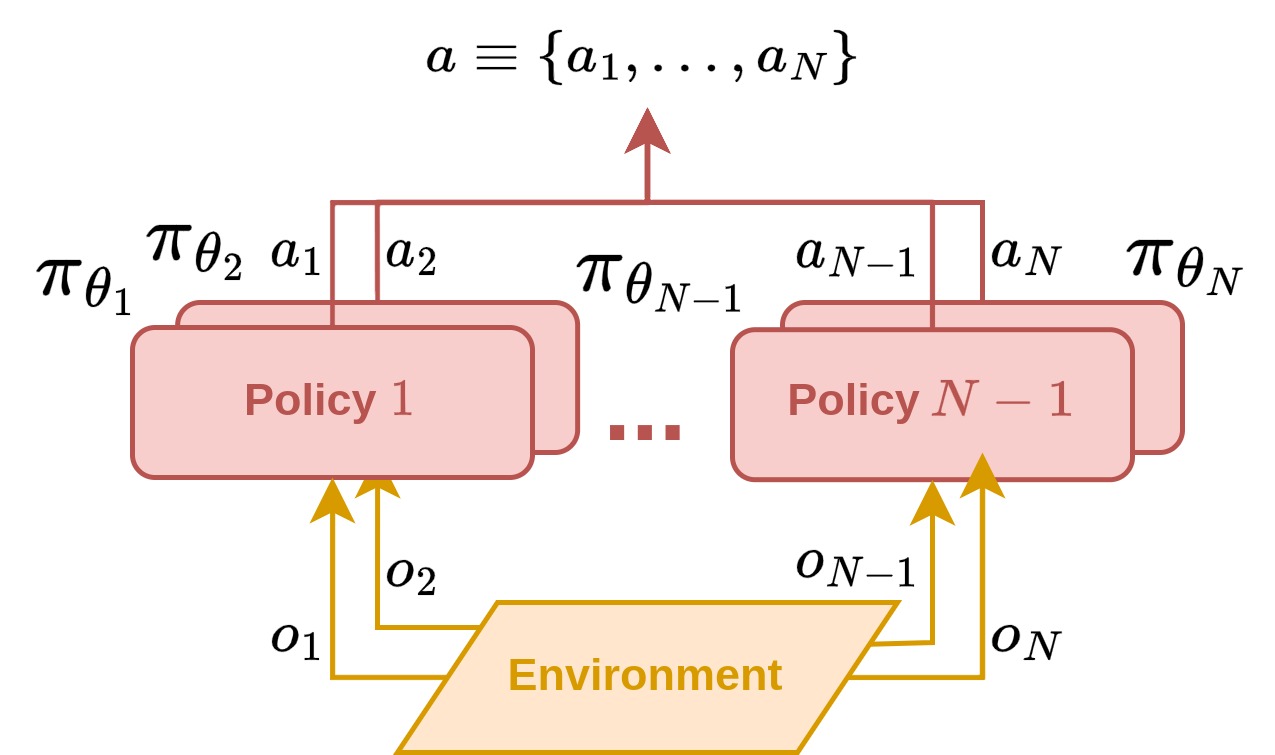}}
    \caption{Simple overview of the configurations of sharing (left) and not sharing (right) parameters. In the former, a joint action is produced by policies that are part of the same network (shared by all the agents), while in the latter each agent has its own separate policy.}
    \label{fig:param_share_vs_no_param_share}
\end{figure}

\subsection{Sharing Parameters and Not Sharing Parameters}
Parameter sharing is a popular strategy adopted by most MARL approaches that allows all the agents of a team to share the same learning networks \cite{gupta_2017}. One key aspect of this approach is the use of an agent ID together with each agent’s input when these are fed to the network. As stated and supported by multiple previous works, this trick will allow the same network to treat the agents as independent units, despite the inputs of all agents being given at the same time and to the same network. According to the literature, the ultimate benefit of this strategy is mainly seen in the training time that the agents take to reach convergence in the tasks and hence on the sample efficiency \cite{rashid_qmix_2018,christianos2021scaling}. 

Not using parameter sharing is empirically not beneficial in simulated environments. However, it allows us to have a different perspective on MARL and look into the learning agents as fully self-contained entities (like humans), and that learn by themselves. Importantly, when we consider practical applications, it is often unfeasible to keep a network that is shared by all the agents \cite{wong2023deep}.

In Figure \ref{fig:param_share_vs_no_param_share}, we can see the key differences between the two described configurations. As it is illustrated, in the parameter sharing setting, all the policies are controlled by the same network with parameters $\theta$, producing a joint action that contains all the actions of the agents. Instead, with the non-parameter sharing configuration, each agent is controlled by an independent network with parameters $\theta_i$ that produces the action only for this particular agent. After all the actions are computed, these are put together into a joint action that is executed in the environment.
\begin{figure}[!t]
    \centering 
    \subfigure[Communication with Parameter Sharing (PS)]{\label{fig:comm_ps_vs_comm_nps_a}\includegraphics[width=0.48\columnwidth]{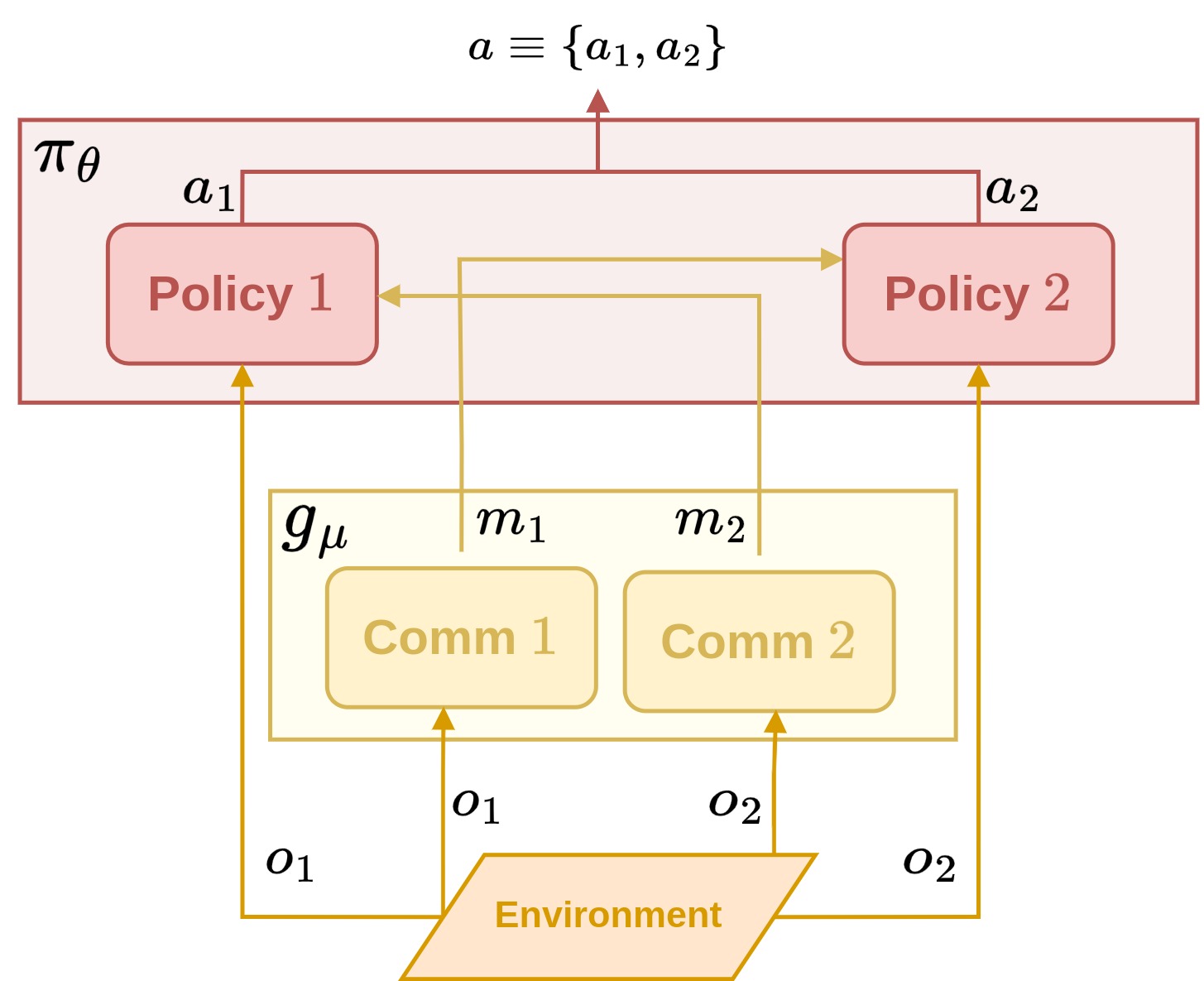}}
    \hfill
    \subfigure[Communication with No Parameter Sharing (NPS)]{\label{fig:comm_ps_vs_comm_nps_b}\includegraphics[width=0.48\columnwidth]{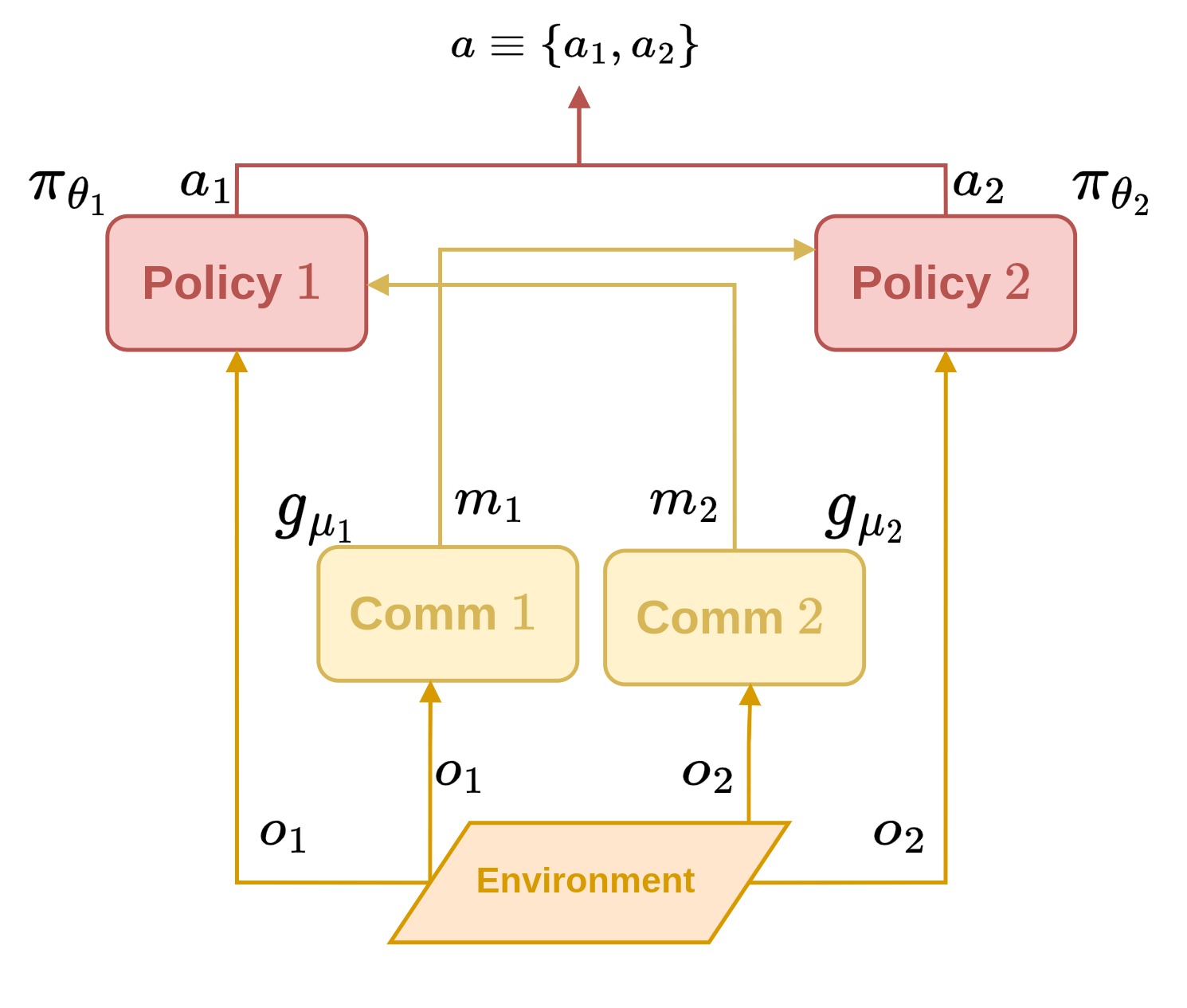}}
    \caption{Illustration of the main differences in the process of generating and broadcasting messages between sharing and not sharing parameters of the learning networks. In the first case, both the policies and communication networks are controlled by the same parameters $\theta$ and $\mu$, while in the second case, these have distinct parameters $\theta_i$ and $\mu_i$.}
    \label{fig:comm_ps_vs_comm_nps}
\end{figure}

\subsection{Communication in MARL}
As it has been discussed throughout this paper, communication in MARL consists on the ability of the agents to share some of their experience of the environment with the teammates. This can include elements such as the observations, the actions, and even a fingerprint. Thus, by using this additional layer of information in the learning process, the agents will learn a Q-function that conditions not only on the action-observation history as in traditional approaches but also in a set of incoming messages from the others, $Q_i([\tau_i, m_{-i}], a_i)$. 

Figure \ref{fig:comm_ps_vs_comm_nps} depicts an overview of the message generation process in MARL. In Figure \ref{fig:comm_ps_vs_comm_nps_a}, we can see that, when the agents share parameters of the learning networks, the process is relatively straightforward: a communication network controlled by the parameters $\mu$ will generate the messages (in this case an encoding of the observations), and then these are sent to the other agents and fed to their policy network together with the observations. Figure \ref{fig:comm_ps_vs_comm_nps_b} now depicts the communication process but when parameters are not shared. Logically, in this case, each agent controls independent networks both for the policy controlled by $\theta_i$ and for communication controlled by $\mu_i$. In this case, each agent independently generates a message that is then sent to the others. While this might seem simple at first, it breaks the link in the computational graph for the communication networks when the losses are backpropagated, since the parameters of the networks are not shared and the final Q-values that result from the policies are conditioned only on the incoming messages. We discuss this phenomenon further ahead. 
\begin{figure*}[!t]
    \centering
    \includegraphics[width=\textwidth]{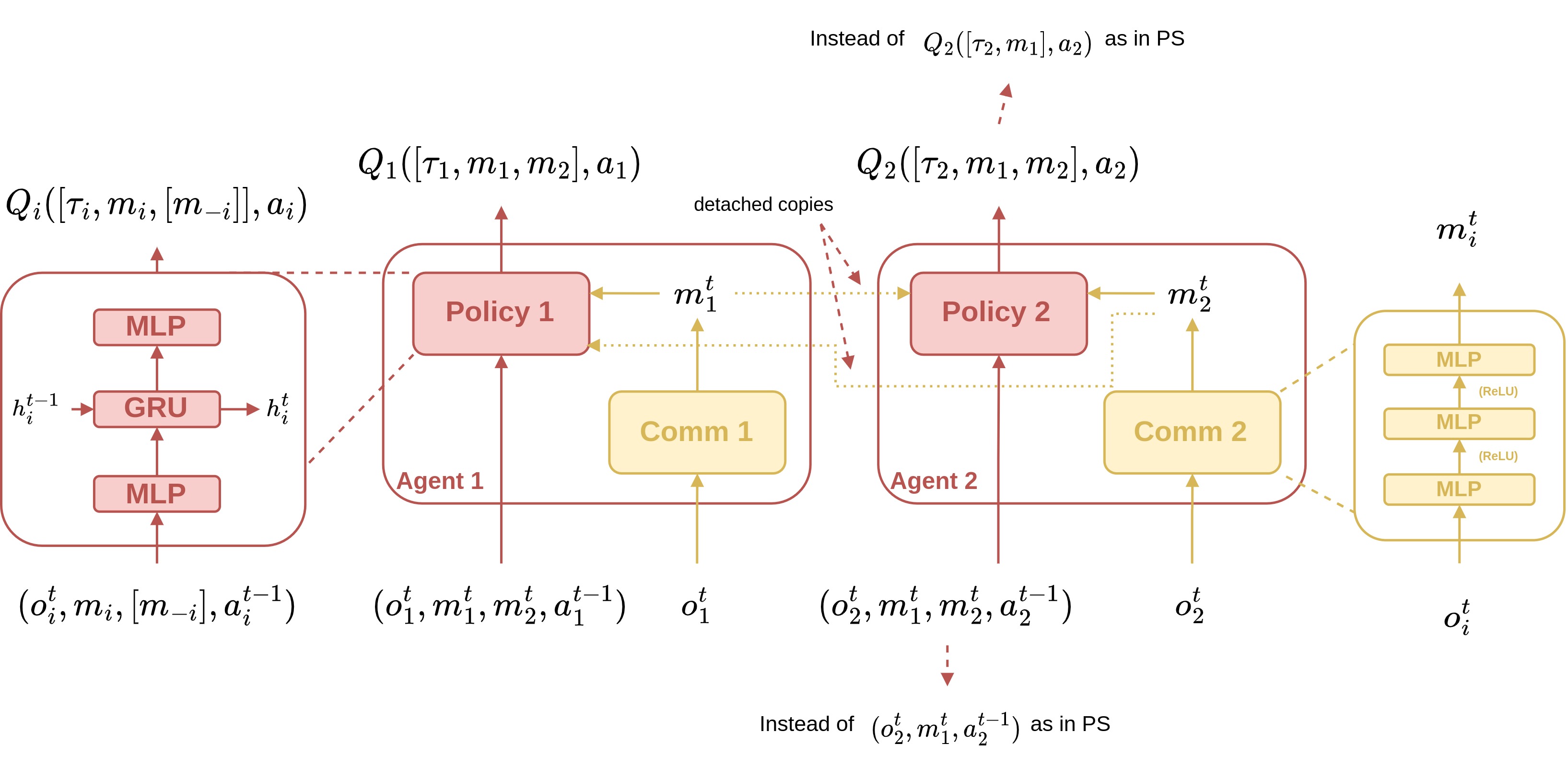}
    \caption{Illustration of how our proposed scheme for independent communication without parameter sharing (NPS+IQL+COMM) works when compared to sharing parameters (PS). The figure shows that agents that do not share parameters also need to receive their own message as input to keep the link to the computational graph of their communication network during backpropagation. On the other hand, when parameters are shared this trick is not needed since all of them use the same network and there are no gradient propagation problems by losing the links to the communication networks in the computation graph.}
    \label{fig:iql_comm_arch}
\end{figure*}

\section{Methods}
\label{sec:setup_and_meths}
\subsection{Communication in MARL for Fully Independent Learners}\label{sec:indep_comm_meth}
In this section, we start by formally describing the implications of communication with fully independent agents in MARL that do not share parameters. Sharing parameters is taken for granted in most MARL approaches, and thus it is often forgotten to consider the implications of not sharing parameters. Importantly, while sharing parameters can be easily done in simulated environments, it is something very difficult to achieve in practical scenarios \cite{wong2023deep}. Thus, it is important to study the implications of not sharing parameters. 

In this paper, we intend to analyse how communication can be integrated with independent learners that do not share parameters. In other words, these agents have their own networks and must learn how to generate messages without receiving direct feedback from the policies of the other agents. In \cite{foerster_learning_2016}, despite they have not shown it, the authors discuss how RIAL could be extended to the case where parameters are not shared. In the case of \cite{foerster_learning_2016} it would be simple since the messages are treated as if they were actions that are sent, and that are generated from the same policy network, leading to lower quality messages \cite{bhalla_training_coop_ag_2019,houidi_2023}. Thus, there are no problems when the losses are propagated in the network because the estimated Q-values ensure that the links to the policy networks are kept. However, training a policy and learning messages at the same time using the same network might represent a difficult learning trade-off \cite{ossenkopf_hierarch_coord_2019}. If the agents have instead a policy and a distinct network specialised for communication they can learn stronger behaviours, but the process becomes challenging. We now formally describe this process and highlight the main problems that need to be addressed.

We consider agents that are controlled by a standard policy network and also have a distinct network whose purpose is to generate messages for communication. Let us first consider the case of IQL \textbf{with} parameter sharing and communication. For simplicity of notation, in the demonstration, we use the observations $o_i$ instead of the history $\tau_i$. Let also $f_i$ and $g_i$ denote two certain functions such that $f_i\rightarrow Q$ and $g_i\rightarrow \mathcal{M}$, for a set of all Q-values $Q$ and a set of all messages $\mathcal{M}$. We have that
\begin{equation}
    \{Q_i\}_{i=1}^N=\{f_i(o_i, m_{-i}, a_i;\theta)\}_{i=1}^N,
\end{equation}
where $m_{-i}$ corresponds to the messages from all agents except $i$, that is produced by a neural network denoted by a function $g_j$ with parameters $\mu$
\begin{equation}
    m_{-i}=\{g_j(o_j;\mu)\}_{j=1,j\neq i}^N\wedge m_i=g_i(o_i;\mu).
\end{equation}

As per Eq. (\ref{eq:vff_loss}), we can define the loss function for the learning problem as
\begin{align}
    \mathcal{L}_i(\theta, \mu)&=r+\gamma max_{a_i'}Q_i(o_i',m_{-i}', a_i';\theta^-)-Q_i(o_i,m_{-i},a_i;\theta)\nonumber\\
    &=r+\gamma max_{a_i'}f_i(o_i',\{g_j(o_j';\mu^-)\}_{j=1,j\neq i}^{j=N},a_i';\theta^-)\nonumber\\
    &-f_i(o_i,\{g_j(o_j;\mu)\}_{j=1,j\neq i}^{j=N},a_i;\theta).
\end{align}
From the above, we can write $\mathcal{L}_i(\theta,\mu)\equiv \mathcal{L}_i(f_i(\cdot;\theta,\mu),g_i(\cdot;\mu))$, and then we can also write the backpropagation rules for the gradients as
\begin{equation}
    \nabla_{\theta}\mathcal{L}_i=\frac{\partial \mathcal{L}_i}{\partial f_i}\frac{\partial f_i}{\partial \theta}+\frac{\partial \mathcal{L}_i}{\partial g_i}\frac{\partial g_i}{\partial \theta}=\frac{\partial \mathcal{L}_i}{\partial f_i} \frac{\partial f_i}{\partial \theta},
\end{equation}
\begin{equation}
    \nabla_{\mu}\mathcal{L}_i=\frac{\partial \mathcal{L}_i}{\partial f_i}  \frac{\partial f_i}{\partial \mu}+\frac{\partial \mathcal{L}_i}{\partial g_i}  \frac{\partial g_i}{\partial \mu},
\end{equation}
and from this, it follows that the parameters of the networks are updated as
\begin{equation}\label{eq:param_up_1}
    \theta = \theta-\alpha\nabla_{\theta}\mathcal{L}_i=\theta-\alpha\frac{\partial \mathcal{L}_i}{\partial f_i}  \frac{\partial f_i}{\partial \theta},
\end{equation}
\begin{equation}
    \mu=\mu-\alpha\nabla_{\mu}\mathcal{L}_i=\mu-\alpha\left(\frac{\partial\mathcal{L}_i}{\partial f_i} \frac{\partial f_i}{\partial\mu}+\frac{\partial\mathcal{L}_i}{\partial g_i} \frac{\partial g_i}{\partial\mu}\right).
\end{equation}

This is the standard procedure for IQL \textbf{with} parameter sharing. However, when we do not share parameters of the networks, the case can be very different. We consider now the setting of fully independent learners, i.e., IQL with \textbf{no} parameter sharing and with communication. In this configuration, the agents are fully self-contained and do not share any parameters. However, we allow them to communicate. In this case, if we follow an equivalent communication scheme as in the previous case (i.e., learning from the incoming messages from the others), we now have that
\begin{equation}
    \{Q_i\}_{i=1}^N=\{f_i(o_i, m_{-i}, a_i;\theta_i)\}_{i=1}^N,
\end{equation}
where $m_{-i}$ corresponds once again to the messages from all agents except $i$, that are produced by a neural network denoted by a function $g_j$ with parameters $\mu_j$
\begin{equation}\label{eq:msgs_eq_2}
    m_{-i}=\{g_j(o_j;\mu_j)\}_{j=1,j\neq i}^N\wedge m_i=g_i(o_i;\mu_i).
\end{equation}

Similarly to Eq. (\ref{eq:vff_loss}), we can define the loss function for the learning problem as
\begin{align}
    \mathcal{L}(\theta_i, \mu_{-i})&=r+\gamma max_{a_i'}Q_i(o_i',m_{-i}', a_i';\theta_i^-)-Q_i(o_i,m_{-i},a_i;\theta_i)\nonumber\\
    &=r+\gamma max_{a_i'}f_i(o_i',\{g_j(o_j';\mu_j^-)\}_{j=1,j\neq i}^{j=N},a_i';\theta_i^-)\nonumber\\
    &-f_i(o_i,\{g_j(o_j;\mu_j)\}_{j=1,j\neq i}^{j=N},a_i;\theta_i).
\end{align}
Because now networks are not shared, from the above we can write $\mathcal{L}_i(\theta_i,\mu_{-i})\equiv \mathcal{L}_i(f_i(\cdot;\theta_i,\mu_{-i}),g_{-i}(\cdot;\mu_{-i}))$.
Thus, hypothetically, in order to update $\mu_i$ the corresponding gradient rule in this case would have to be
\begin{equation}
    \nabla_{\mu_i}\mathcal{L}_j=\frac{\partial\mathcal{L}_j}{\partial f_j} \frac{\partial f_j}{\partial\mu_i}+\frac{\partial\mathcal{L}_j}{\partial g_{-j}} \frac{\partial g_{-j}}{\mu_i},
\end{equation}
which is an absurd, because $j$ does not share parameters with $i$, and thus $\mu_i$ will never be updated $\forall i \in \{1,\dots,N\}$. This can be summarised as the following remark.

\begin{remark}\label{rem:rem_1}
    The parameters of a communication network $\mu_i$ of agent $i$ will never be updated if fully independent learners that do not share the parameters $\theta_i$ and $\mu_i$ learn only from their observations and incoming messages from the others. 
\end{remark}

As a solution to the problem stated in Remark \ref{rem:rem_1} that occurs in independent communication without sharing parameters when updating the communication networks, we propose instead the following learning scheme for independent communication:
\begin{equation}\label{eq:q_eq_3}
    \{Q_i\}_{i=1}^N=\{f_i(o_i, m_{-i},m_i,a_i;\theta_i)\}_{i=1}^N,
\end{equation}
where $m_{-i}$ corresponds once again to the messages from all agents except $i$, that are produced by a function $g_j$ with parameters $\mu_j$, in the same way as in Eq. (\ref{eq:msgs_eq_2}). $\mathcal{L}_i$ can now be written as
\begin{align}
    \mathcal{L}_i(&\theta_i, \mu_{-i}, \mu_i)=\nonumber\\
    &=r+\gamma max_{a_i'}Q_i(o_i',m_{-i}',m_i',a_i';\theta_i^-)-Q_i(o_i,m_{-i},m_i,a_i;\theta_i)\nonumber\\
    &=r+\gamma max_{a_i'}f_i(o_i',\{g_j(o_j';\mu_j^-)\}_{j=1,j\neq i}^{j=N},g_i(o_i';\mu_i^-),a_i';\theta_i^-)\nonumber\\
    &-f_i(o_i,\{g_j(o_j;\mu_j)\}_{j=1,j\neq i}^{j=N},g_i(o_i;\mu_i),a_i;\theta_i),
\end{align}
\sloppy and now, we have that $\mathcal{L}_i(\theta_i,\mu_{-i},\mu_i)\equiv\mathcal{L}_i(f_i(\cdot;\theta_i,\mu_{-i},\mu_i),g_{-i}(\cdot;\mu_{-i}),g_i(\cdot;\mu_i))$, and we can now write the rules as
\begin{equation}\label{eq:mu_partial_15}
    \nabla_{\theta_i}\mathcal{L}_i=\frac{\partial \mathcal{L}_i}{\partial f_i} \frac{\partial f_i}{\partial \theta_i}+\frac{\partial\mathcal{L}_i}{\partial g_{-i}} \frac{\partial g_{-i}}{\partial \theta_i}+\frac{\partial\mathcal{L}_i}{\partial g_i} \frac{\partial g_i}{\partial\theta_i}=\frac{\partial\mathcal{L}_i}{\partial f_i} \frac{\partial f_i}{\partial \theta_i},
\end{equation}
\begin{equation}\label{eq:mu_partial_16}
    \nabla_{\mu_i}\mathcal{L}_i=\frac{\partial \mathcal{L}_i}{\partial f_i} \frac{\partial f_i}{\partial \mu_i}+\frac{\partial\mathcal{L}_i}{\partial g_{-i}} \frac{\partial g_{-i}}{\partial \mu_i}+\frac{\partial\mathcal{L}_i}{\partial g_i} \frac{\partial g_i}{\partial\mu_i}=\frac{\partial\mathcal{L}_i}{\partial f_i} \frac{\partial f_i}{\partial \mu_i}+\frac{\partial\mathcal{L}_i}{\partial g_i} \frac{\partial g_i}{\partial\mu_i}.
\end{equation}
Intuitively, this step solves the problem stated in Remark \ref{rem:rem_1}. However, now when doing the final rule, it is implied that
\begin{align}\label{eq:mu_partial_17}
    \nabla_{\mu_{-i}}\mathcal{L}_i&=\frac{\partial\mathcal{L}_i}{\partial f_i} \frac{\partial f_i}{\partial\mu_{-i}}+\frac{\partial\mathcal{L}_i}{\partial g_{-i}} \frac{\partial g_{-i}}{\partial\mu_{-i}}+\frac{\partial\mathcal{L}_i}{\partial g_i} \frac{\partial g_i}{\partial\mu_{-i}}\nonumber\\
    &=\frac{\partial\mathcal{L}_i}{\partial f_i} \frac{\partial f_i}{\partial\mu_{-i}}+\frac{\partial\mathcal{L}_i}{\partial g_{-i}} \frac{\partial g_{-i}}{\partial\mu_{-i}}.
\end{align}
From Eq. (\ref{eq:mu_partial_17}), we note the existence of a second problem (that is independent of our solution to the problem in Remark \ref{rem:rem_1}), since $\mu_{-i}$ would be updated as
\begin{equation}
    \mu_{-i}=\mu_{-i}-\alpha\left(\frac{\partial\mathcal{L}_i}{\partial f_i} \frac{\partial f_i}{\partial\mu_{-i}}+\frac{\partial\mathcal{L}_i}{\partial g_{-i}} \frac{\partial g_{-i}}{\partial\mu_{-i}}\right),
\end{equation}
for $N$ times, causing losses of gradient when propagating through the same values several times. We can write a second important remark

\begin{remark}\label{rem:rem_2}
    If fully independent agents that do not share the parameters $\theta_i$ and $\mu_i$ of their networks learn from the messages of the others, then the incoming messages will be used for backpropagation $N$ times, causing problems in the computational graph. 
\end{remark}
To overcome the problem described in Remark \ref{rem:rem_2}, for each agent $i$, we detach $m_{-i}$ from the computational graph, ensuring that all $\theta_i \wedge \mu_i, i\in\{1,\dots,N\}$ are updated exactly once, according to
\begin{equation}\label{eq:param_up_2}
    \theta_i=\theta_i-\alpha\frac{\partial \mathcal{L}_i}{\partial f_i} \frac{\partial f_i}{\partial\theta_i},\quad\text{as per Eq. (\ref{eq:mu_partial_15})},
\end{equation}
\begin{equation}
    \mu_i=\mu_i-\alpha\left(\frac{\partial \mathcal{L}_i}{\partial f_i} \frac{\partial f_i}{\partial \mu_i}+\frac{\partial \mathcal{L}_i}{\partial g_i} \frac{\partial g_i}{\partial \mu_i}\right),\quad\text{as per Eq. (\ref{eq:mu_partial_16})}.
\end{equation}

With this learning scheme, which can be summarized by Eq. (\ref{eq:q_eq_3}), we solve both problems described in Remarks \ref{rem:rem_1} and \ref{rem:rem_2} that occur in fully independent learning with communication and without parameter sharing. This scheme allows all the parameters to be updated, enabling learning with communication. In the results section ahead, we show that the agents that follow this configuration are still able to learn communication strategies.
\begin{algorithm}[!t] 
    \begin{algorithmic}[1]
    \setstretch{0.9}
    \State Initialise empty replay buffer $\mathcal{D}$ and the parameters of policy and communication networks for each agent $i$
    \For {step = 0 to maximum steps}
      \While {episode is not done}
        \State Execute action $a=(a_1,\dots,a_N)$ and get $r$ and $s'$
        \State Update buffer $\mathcal{D}$ with $(s,a,r,s')$
    \EndWhile
    \If {$\mathcal{D}$ is not empty}
      \For {each episode e in $B\sim\mathcal{D}$}
          \For {each agent i} \Comment{Message generation}
              \State Send $m_i \gets comm\_network(\tau_i)$ to others $\neq i$
          \EndFor
          \For {each agent i} \Comment{Training stage}
              \State $m_{-i} \gets$ incoming messages from others $\neq i$
              \State $m_{-i}^d \gets$ detached copy of $m_{-i}$
              \State $Q_i\equiv Q_i([\tau_i,m_{-i}^d,m_i],a_i)$
              \State Calculate targets $Q_i'$ using target networks
              \State $\Delta Q_i=r+\gamma max_{a_i'}Q_i' - Q_i$
              \State Update all $\theta_i$ parameters
              \State Update target networks every target interval
          \EndFor
      \EndFor
    \EndIf
    \EndFor
    \end{algorithmic} 
    \caption{Fully Independent Communication}
    \label{alg:c6_algorithm2}
\end{algorithm}
  
\subsection{Setting}
To conduct the experiments in this paper, we consider a set of different algorithmic configurations to enable the analysis of the effects of communication with and without parameter sharing. We opt to use always independent Q-learning (IQL with deep Q-networks, as described in section \ref{sec:bkg_iql}, due to its known simplicity, allowing for a fair analysis of the implications of different configurations involving communication and varying levels of information exchange. In addition, it facilitates the analysis of the impact of sharing and not sharing parameters. Note that the scope of this paper is not to propose complex communication architectures, as in other works such as \cite{ding2022sequential,liu_multi-agent_2021,kim2019learning}. We consider the following configurations: 
\begin{itemize}
    \item PS+IQL: refers to the use of IQL and the agents share the parameters of the same network (parameter sharing). 
    \item NPS+IQL: refers to the use of IQL and the agents do not share the parameters of their networks (\textbf{no} parameter sharing). 
    \item PS+IQL+COMM: refers to the use of IQL with the agents sharing parameters (same as PS+IQL), but now we include a communication module that allows the agents to broadcast messages; we provide more details ahead. 
    \item NPS+IQL+COMM (that corresponds to the proposed learning scheme in \ref{sec:indep_comm_meth}): refers to the use of IQL with the agents \textbf{not} sharing parameters (same as NPS+IQL), but now they use a communication module (same as in PS+IQL+COMM) that allows the agents to broadcast messages; note that here, since we do not consider parameter sharing, the communication should also be self-contained and each agent should encode its own messages (with its own communication network). Algorithm \ref{alg:c6_algorithm2} describes how this method works.
\end{itemize}

Figure \ref{fig:iql_comm_arch} shows the general architecture of the communication approaches. The figure depicts directly the architecture of NPS+IQL+COMM (as described in subsection \ref{sec:indep_comm_meth}). For brevity, we do not show the architectures of the other configurations, although these can be deduced directly from this figure. If the communication modules are removed, then it represents NPS+IQL, and then both configurations can be extended to the PS case if we assume the networks to be representations of the same network. Nonetheless, in the supplementary, we still include the architecture without communication.
\begin{figure}[!t]
    \centering 
    \subfigure[3s\_vs\_5z]{\label{fig:env_ss_smac}\includegraphics[width=0.48\columnwidth]{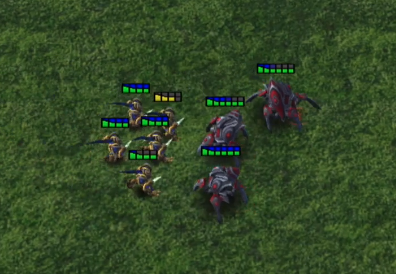}}
    \hfill
    \subfigure[PredatorPrey]{\label{fig:env_ss_pp}\includegraphics[width=0.35\columnwidth]{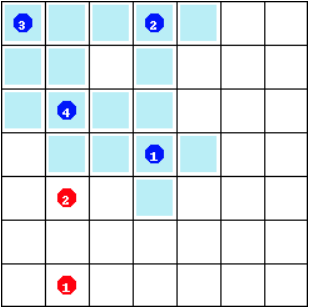}}
    \caption{Environments used in the experiments. On the left, 3s\_vs\_5z, a scenario from the SMAC collection \cite{smac_2019}, and on the right a PredatorPrey game, where 4 predators must catch two moving prey \cite{magym}.}
    \label{fig:envs_ss}
\end{figure}

\subsection{Network Capacity}   
In this paper, to carry out a deeper analysis of how communication affects learning, we also investigate how changing the dimension number of the hidden layers of the agents’ networks affects their learning process in the presence of communication. When we increase the size of the hidden layers, it means that the policy network can represent more information than when this size is decreased. This means that, if agents can learn using networks that have lower capacity, these agents can still perform under lightweight networks. Importantly, these networks can still extract the needed information from the inputs to the network to learn the task, while keeping the representations of the inputs much smaller. This can be crucial when there are certain computational constraints. At the same time, it means that lots of capacity might be useless and there is waste of information if the same information could be represented with a lower capacity.  

\section{Experiments and Results\protect\footnote{C\MakeLowercase{odes used can be found at \href{https://github.com/rafaelmp2/marl-indep-comm}{https://github.com/rafaelmp2/marl-indep-comm}}}}\label{sec:exps_and_res}
In this section, we present the results of the discussed different configurations. Importantly, one of the key points of these experiments is to evaluate whether the proposed method in section \ref{sec:indep_comm_meth} enables successful communication for independent learners who do not share parameters. We evaluate our hypotheses in the environments 3s\_vs\_5z and PredatorPrey (Figures \ref{fig:env_ss_smac} and \ref{fig:env_ss_pp}). 3s\_vs\_5z is one of the environments of the SMAC collection \cite{smac_2019}, where 3 stalker units (melee units) must defeat 5 zealots (ranged units). In the used version of PredatorPrey, 4 agents must capture 2 randomly moving prey in a $7\times7$ grid world. The team reward that is received is $5\times N$ when a prey is caught, and there is a step penalty of $-0.1\times N$. Most importantly, there is a team punishment for non-cooperative behaviours of $-0.75\times N$ every time an agent attempts to capture a prey alone. This punishment in PredatorPrey tasks has shown to be important in previous works to evaluate the importance of communication approaches \cite{liu_multi-agent_2021,deep_coord_2019,kim2019learning,zhang2019efficient}, since these punishments will make the task confusing for the agents.

With the results from the experiments carried out in this paper, we intend to answer the following questions: 
\begin{itemize}
    \item (\textbf{Q1}) How does sharing parameters affect learning when compared to not sharing parameters? 
    \item (\textbf{Q2}) Can agents still communicate when they do not share parameters?
    \item (\textbf{Q3}) Is communication always necessary (since it is often naively applied, bringing useless complexity to the learning networks)?
    \item (\textbf{Q4}) How is communication affected by different sizes of network capacity for when we both share and not share parameters? 
\end{itemize}
\begin{figure}[!t]
    \centering 
    \subfigure[PS+IQL]{\label{fig:r_smac_a}\includegraphics[width=0.48\columnwidth]{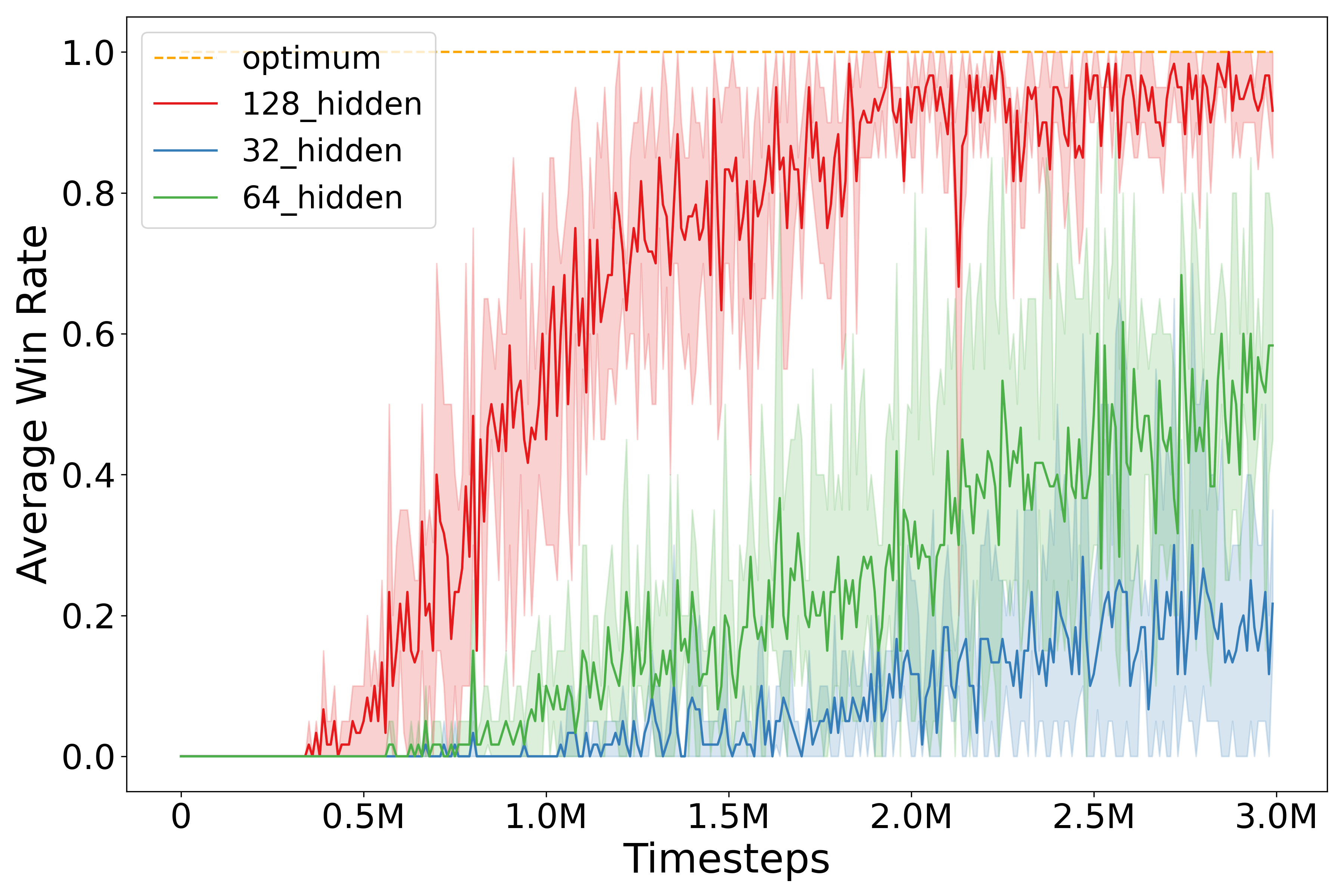}}
    \subfigure[PS+IQL+COMM]{\label{fig:r_smac_b}\includegraphics[width=0.48\columnwidth]{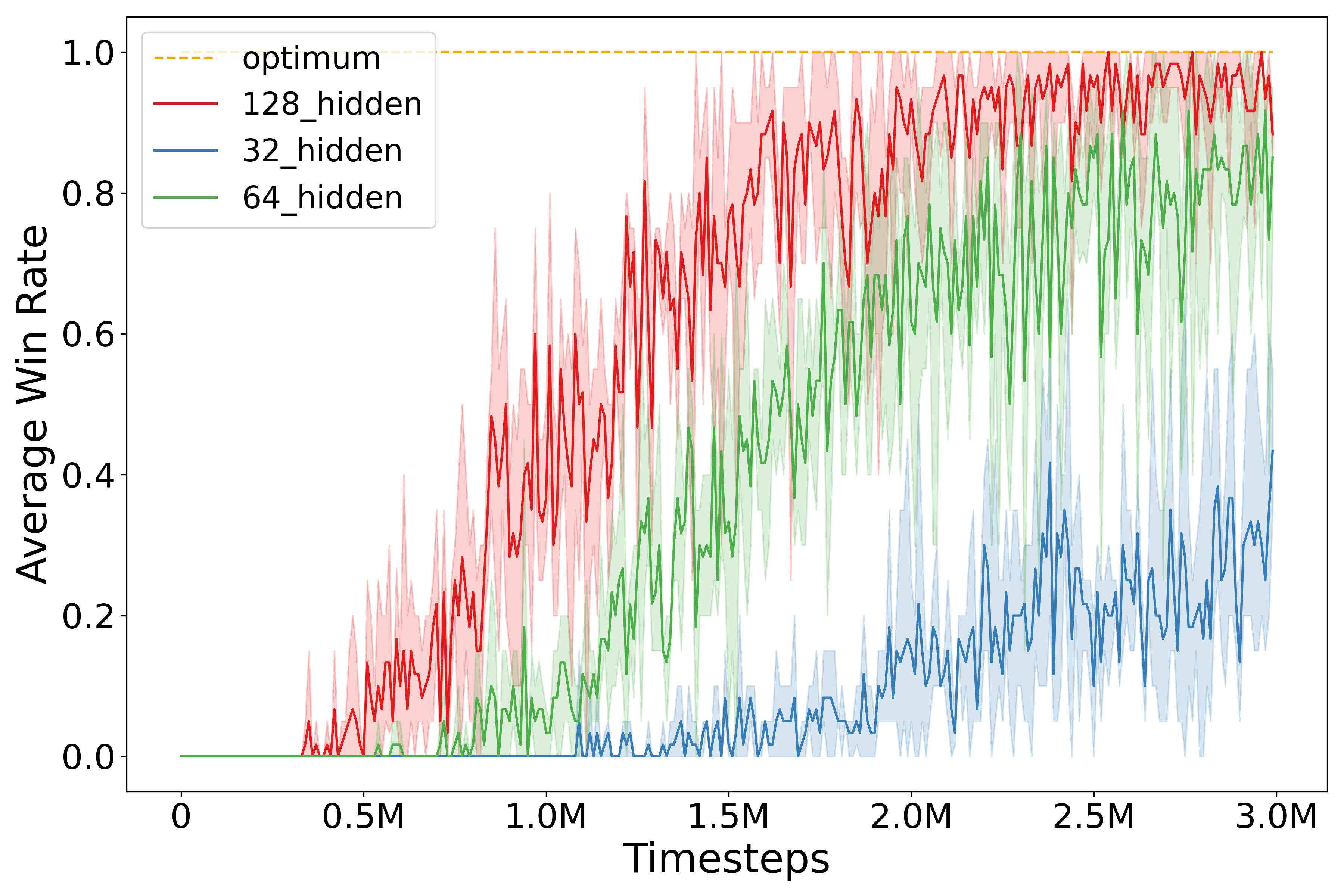}}
    \subfigure[NPS+IQL]{\label{fig:r_smac_c}\includegraphics[width=0.48\columnwidth]{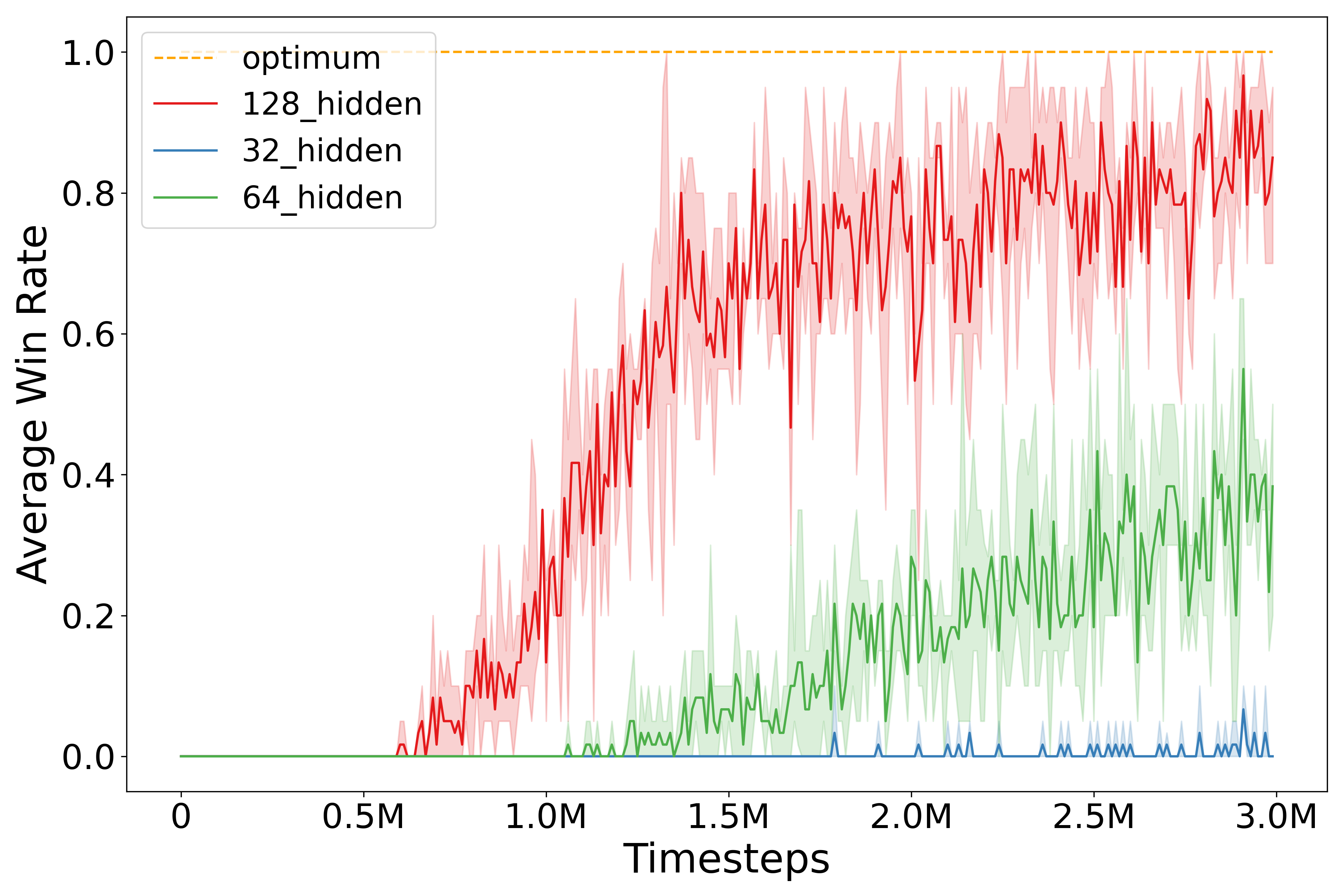}}
    \subfigure[NPS+IQL+COMM]{\label{fig:r_smac_d}\includegraphics[width=0.48\columnwidth]{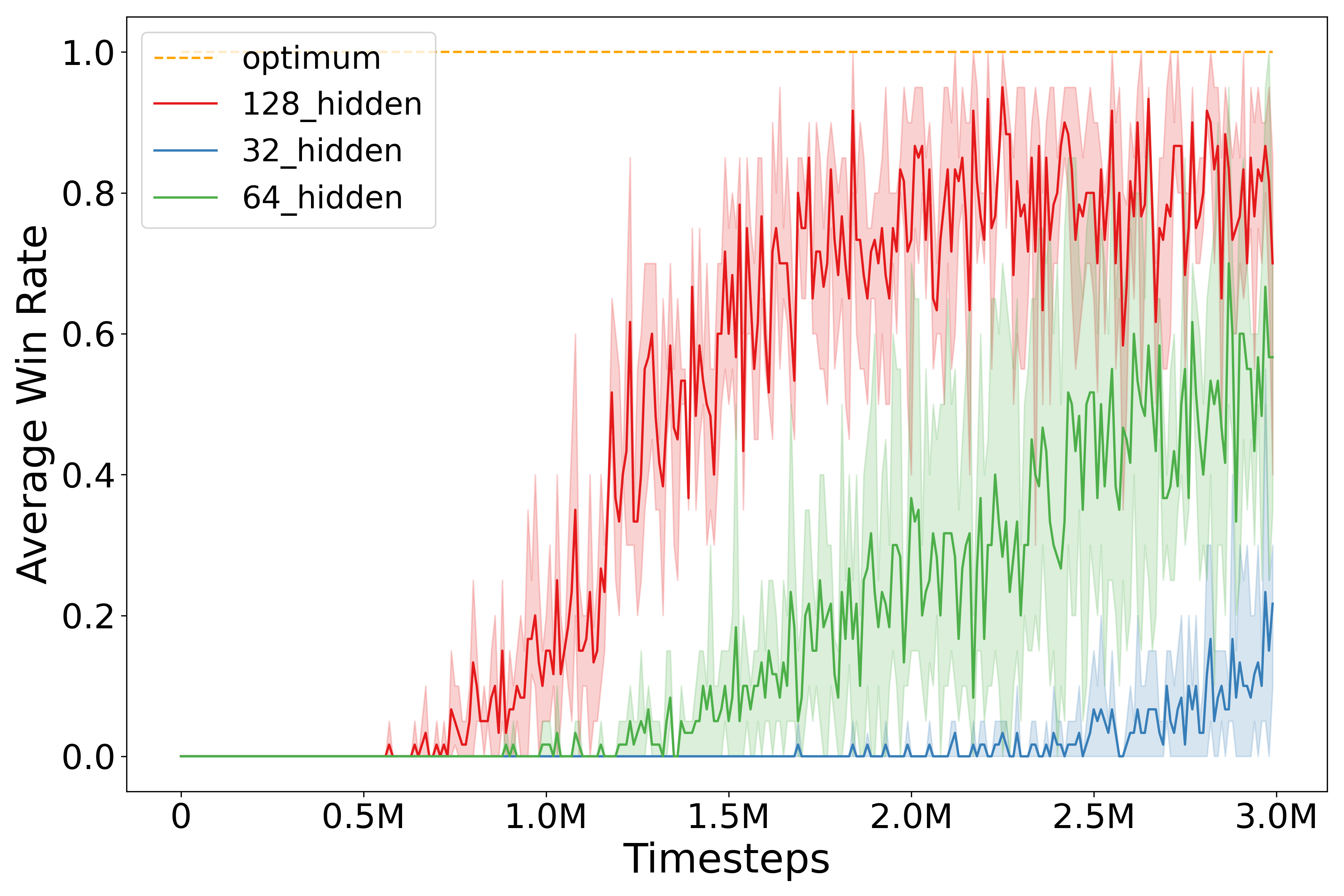}}
    \caption{Win rates achieved by the attempted methods in 3s\_vs\_5z. The dashed line (optimum) represents the optimal value that could be achieved by the agents, i.e., a win rate of 1. For completeness, we include, in the supplementary, the corresponding rewards of these win rates.}
    \label{fig:rewards_3svs5z}
\end{figure}

In the supplementary, we describe the hyperparameters used in the experiments, alongside other implementation details.

\subsection{Communication when Sharing or Not Sharing Parameters}
\begin{figure}[!t]
    \centering 
    \subfigure[PS+IQL]{\label{fig:r_pp_a}\includegraphics[width=0.23\textwidth]{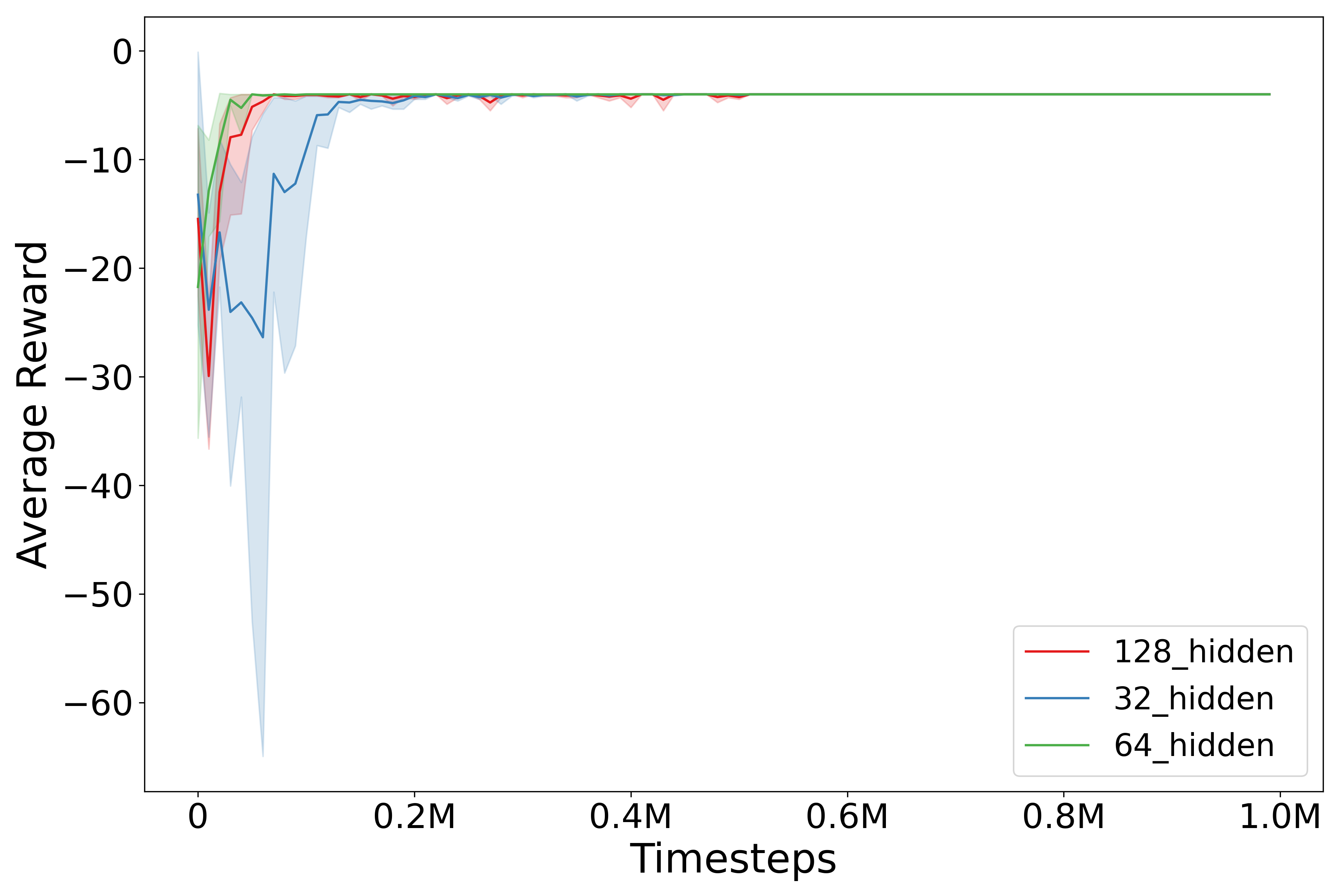}}
    \subfigure[PS+IQL+COMM]{\label{fig:r_pp_b}\includegraphics[width=0.23\textwidth]{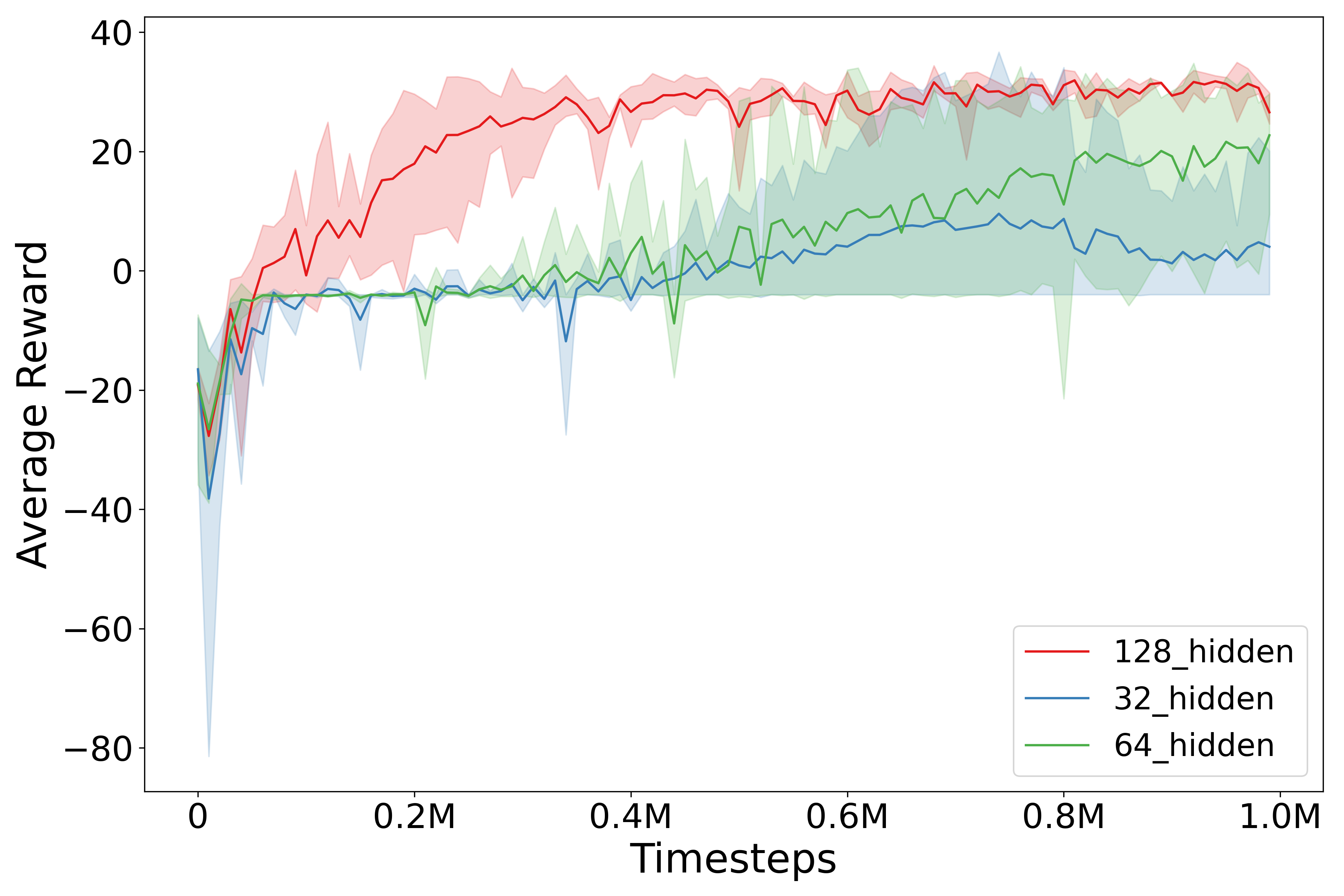}}
    \subfigure[NPS+IQL]{\label{fig:r_pp_c}\includegraphics[width=0.23\textwidth]{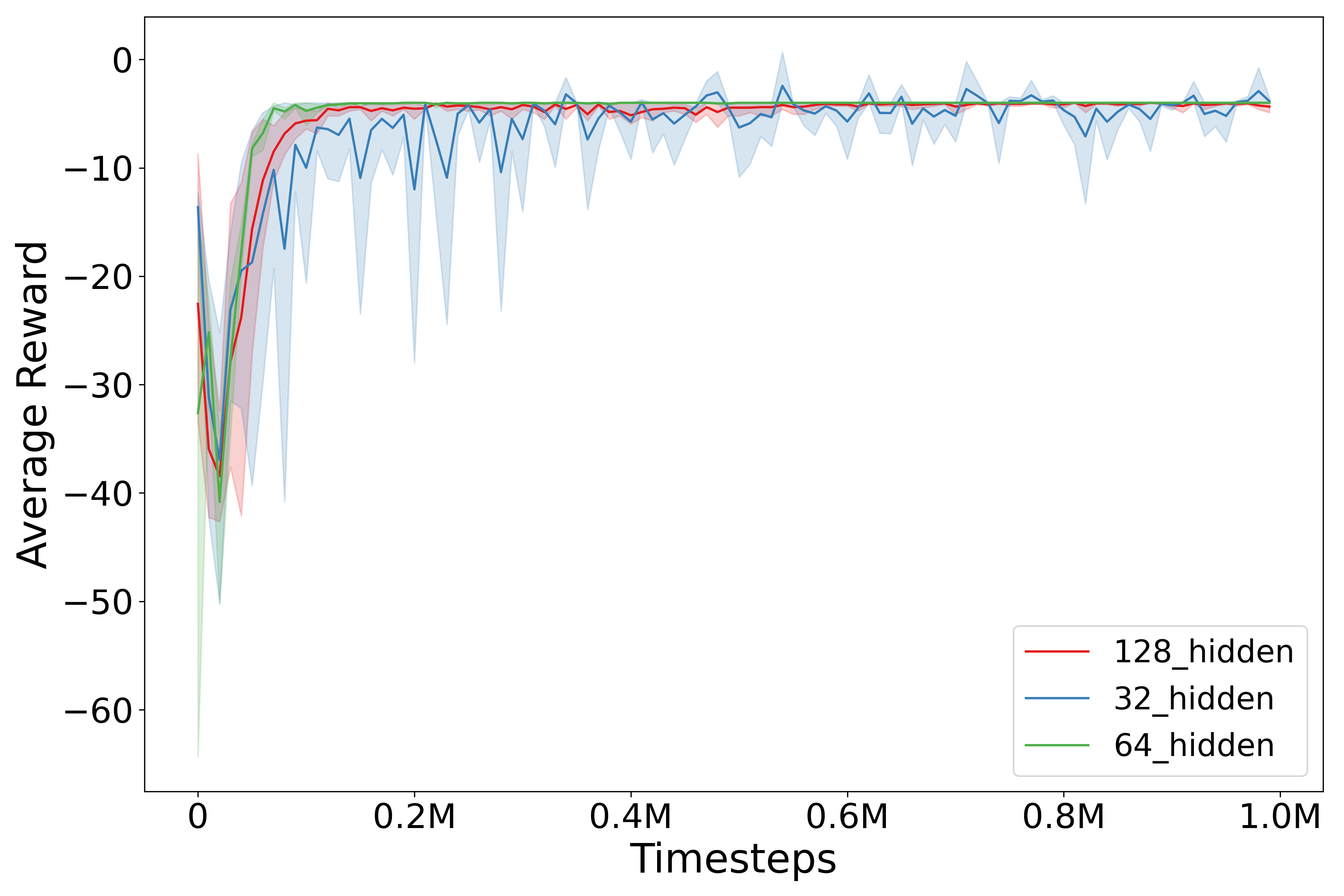}}
    \subfigure[NPS+IQL+COMM]{\label{fig:r_pp_d}\includegraphics[width=0.23\textwidth]{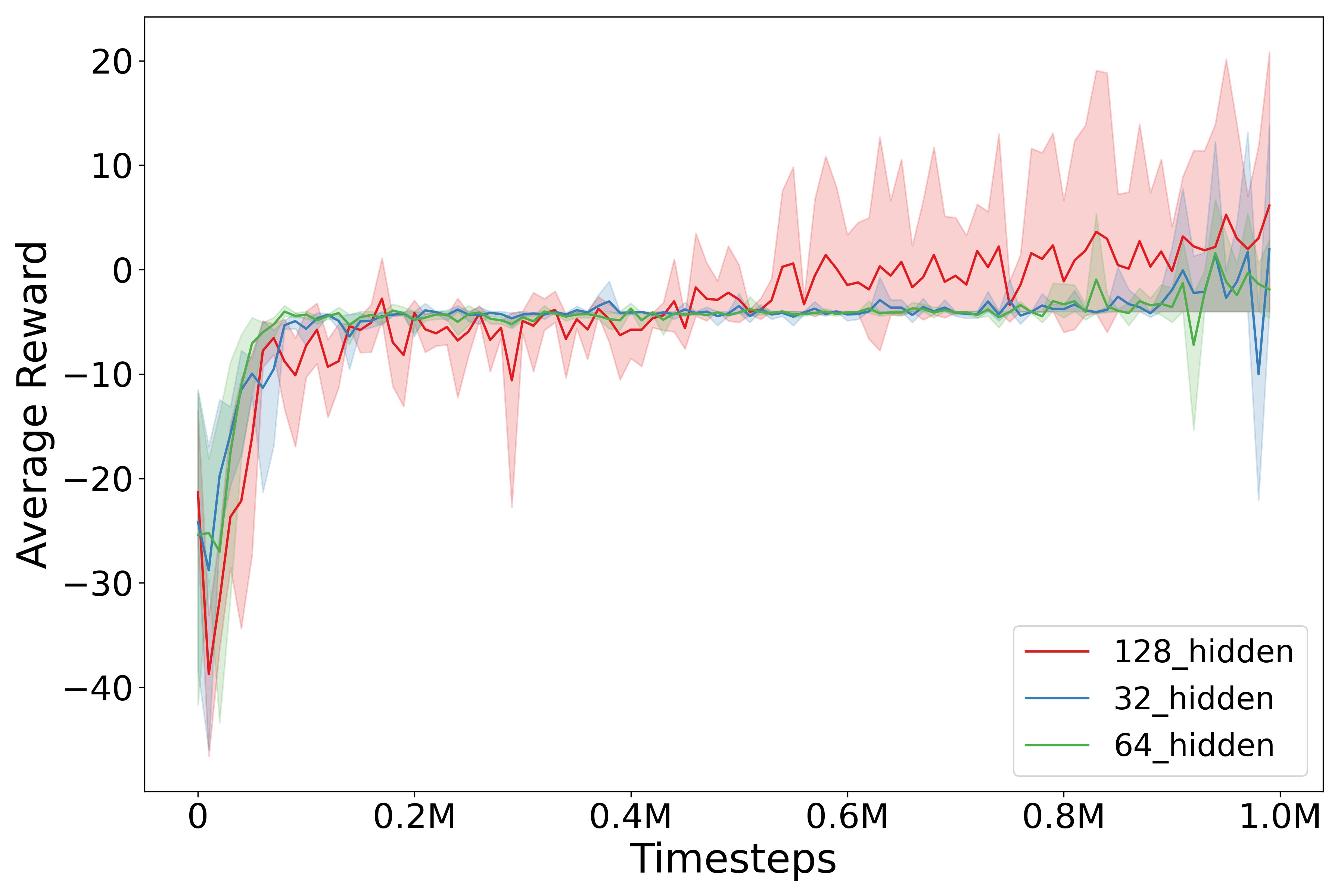}}
    \caption{Rewards achieved by the attempted methods in the PredatorPrey task, with a punishment for non-cooperative behaviours of -0.75.}
    \label{fig:rewards_pp}
\end{figure}

As we have discussed previously in this paper, one of the key points here is to study the implications of sharing and not sharing parameters. From the presented experiments, we observe that sharing parameters naturally brings advantages to the agents' learning process. In line with other works that have argued that sharing parameters works mostly as a way of speeding up learning and saving computational resources, here we enhance in Figure \ref{fig:r_smac_a} that, when the agents share parameters, they solve 3s\_vs\_5z much faster, despite not sharing parameters will also solve the environment, but while taking much longer. When we study the effect of communication, it is possible to see that the impact of communication is also more evident when parameters are shared. This is due to the fact that the feedback resulting from the messages produced and the policies of the agents is backpropagated to the same networks. In the case of not sharing parameters, the problem becomes much more complex because there is no link between networks of different agents, and thus the communication network does not receive direct feedback of the messages produced by itself. Instead, it has to understand the impact of these messages by understanding how they are affecting the team reward globally. However, despite this limitation inherent to the fact that parameters are not shared, we can still see the improvements of communication in fully independent learners in Figure \ref{fig:r_smac_d}. These improvements of communication for fully independent agents are even stronger in PredatorPrey (Figure \ref{fig:r_pp_d}), where the agents manage to achieve positive rewards in the task, as opposed to when they do not communicate (Figure \ref{fig:r_pp_c}). This demonstrates that our framework for communication when parameters are not shared enables learning in this challenging configuration. This is an important observation when we consider scenarios where parameters cannot be shared (\textbf{Q1}, \textbf{Q2}).

It is common to see in the literature on communication in MARL methods where communication is naively applied to an arbitrary number of varied environments. In most cases, the methods using communication end up performing better in the tested environments, but there are cases where the performances achieved are very close to benchmarks without communication being used. In such cases, one often wonders whether communication is really necessary, or if it is just bringing useless overhead to the networks, which can be problematic. Figures \ref{fig:r_smac_a} and \ref{fig:r_smac_b} show the performances of PS+IQL and PS+IQL+COMM in the 3s\_vs\_5z environment, respectively. In this case, we can see that communication does not seem to play an important role in the task. In both cases, the agents achieve optimal performances, meaning that communication becomes redundant and only brings overhead to the agents, since an additional network is being used (and simply increasing the capacity of the agent networks would be enough). When parameters are not shared, NPS+IQL and NPS+IQL+COMM in Figures \ref{fig:r_smac_c} and \ref{fig:r_smac_d}, we observe a similar scenario, where the effect of communication is almost unnoticeable. On the other hand, when we look at the second tested environment PredatorPrey, we can see that communication has a very strong impact (Figure \ref{fig:rewards_pp}). When communication is not used, the agents cannot solve this task at all. However, when they communicate, they can break the barrier of negative returns, both when they share and do not share parameters. Thus, in this case, communication is necessary for learning, as opposed to the previous environment. These results urge for the need to analyse when communication is needed or not before applying it naively as it might cause a waste of computational resources (\textbf{Q3}). 

\subsection{The Impact of Different Network Capacity}
In order to study the amount of network capacity needed for learning and how communication helps with this information, we have also experimented with different sizes for the hidden layers of the agent network but fixed the communication network hidden dimensions to 64. As it was expected, when the agents have a higher network capacity, their performance is drastically improved. On the other hand, when the network capacity is not enough, it might take them longer to learn the tasks. This means that networks with a higher network capacity have a higher sample efficiency as they can learn faster with the same amount of samples. Thus, we can see that, while with size 32 the agents struggle to learn in 3s\_vs\_5z, when we increase to 64 and 128 they learn more easily and much faster. Once again, this verifies both for PS+IQL and NPS+IQL (Figures \ref{fig:r_smac_a} and \ref{fig:r_smac_c}). 

When we look at the communication side (Figures \ref{fig:r_smac_b} and \ref{fig:r_smac_d}), we observe once again that, in the case of 3s\_vs\_5z, following our hypothesis that communication may not be always necessary, adding communication will not yield any significant improvement over simply changing the sizes of the network capacity. However, when we look at the PredatorPrey case (Figure \ref{fig:rewards_pp}), we can see that the agents can only solve it with communication and, while increasing the network capacity without communication does not have any impact, increasing it together with adding communication makes a big difference. Here we can see that communication allied to the right network size will lead to better performance. Thus, by using a communication network, we can spare resources regarding the standard agent networks. In summary, note the important remark that while increasing the network capacity might be enough for some cases, when communication is necessary simply increasing the network capacity is not enough, and both are needed (\textbf{Q4}). 

\section{Related Work}
\label{sec:rel_work}
Recent works in MARL have developed strong methods to tackle complex problems based on neural network architectures that can give good value function estimations. To name a few, QMIX proposes a way of mixing individually estimated action-value functions into a global function, from which optimal joint policies can be extracted \cite{rashid_qmix_2018}. VDN, or QTRAN \cite{sunehag2017valuedecomposition,qtran_2019} are other such methods that learn a different type of mix of these individual functions. Ultimately, the goal is always to learn an optimal joint action-value function. These methods compose a way of learning without communication, i.e., the agents do not directly broadcast information to each other, although they have indirect access to the policies of the others due to the mixing during training.

While these methods are communication-free, lately multiple other works have targeted communication in MARL from different angles. In \cite{foerster_learning_2016} the authors demonstrate how a communication protocol can be learned when the message is generated by the policy of the agents and treated as an action. In \cite{sukhbaatar_learning_2016} it is proposed another method of communication that aggregates the messages of all the agents into a cumulative message that is then sent. In \cite{chu2020multi} the authors propose a method that minimises information loss of message aggregation, but requires access to global information during training. Other recent approaches have proposed strategies for communicating based on specific factors such as whom, when, and what to communicate \cite{das_tarmac_2019,jiang_learning_2018}. In \cite{gupta2023hammer}, the authors use a central agent that controls messages and learns what needs to be sent to the agents. Other types of communication have been used together with methods that do not initially use communication. For instance, methods such as \cite{liu_multi-agent_2021,zhang2019efficient} propose communication architectures that can be used to improve the performance of other non-communicative approaches. Among all the mentioned methods, note that, during the communication process, there is always some sort of information exchange (when sharing messages), and there is always some sort of loss of information (during encoding or aggregation). This type of information conditioning is studied in works such as \cite{wang_bottleneck_2019}, where the authors propose a method that can learn messages that follow certain bandwidth constraints. In \cite{kim2019message}, the authors also use a drop-out strategy based on the average weights of several message networks during training.

Aspects such as the implications of communicating under limited bandwidth channels, or the measurement of the quality of communication have been studied in works such as \cite{resnick_capacity_2020,wang_bottleneck_2019,lowe_pitfalls_2019,tucker_inf_2022}. Instead, in this work we intend to analyse whether communication is really necessary when the agent networks already have enough capacity. In works such as \cite{hu2023rethinking} it is shown how increasing the network capacity can improve learning, but how does it affect communication methods? Furthermore, note that in \cite{vanneste2023distributed} it is analysed communication among critics that approximate advantage utilities. However, the authors do not focus on direct agent-to-agent communication. Importantly, we also note that, while in \cite{foerster_learning_2016} the authors have briefly mentioned communication for independent agents, it is still not clear how communication can be achieved in fully independent agents that do not share parameters in complex scenarios. Sharing parameters is often taken for granted in MARL, leaning research towards oblivion regarding the understanding of the implications of either sharing or not sharing parameters, as discussed in works like \cite{christianos2021scaling}.

\section{Conclusion}
Communication is still an open area of research in MARL. While remarkable progress has been made in the field, there are still aspects that have not been investigated in detail. In this work, we have particularly shown that there are several advantages when agents are allowed to share parameters of the learning networks, as supported by the literature. However, this configuration may not always be feasible (for instance, in practical applications), and thus there is a need to deepen the study of the implications of not sharing parameters. We have proposed a way of communicating in MARL for independent learners that have distinct networks for policy and message generation and do not share parameters. The results achieved show that agents can still learn communication strategies under this setting.

Taking into account the network capacity can also be a deal breaker in learning. Even when considering communication among agents, different network capacities may lead to different outcomes. In fact, it is important to evaluate whether communication or extra network capacity in MARL is really needed before jumping to these options naively, resulting in unnecessary overhead. In the future, we intend to extend our findings to other different scenarios and dig deeper into the learning process of communication that results from the proposed learning scheme for fully independent communication.


\newpage
\appendix
\onecolumn

\section{SMAC Reward Plots}\label{ap:ap_a}
\begin{figure}[!hbt]
    \centering 
    \subfigure[PS+IQL]{\label{fig:app_r_smac_a}\includegraphics[width=0.23\textwidth]{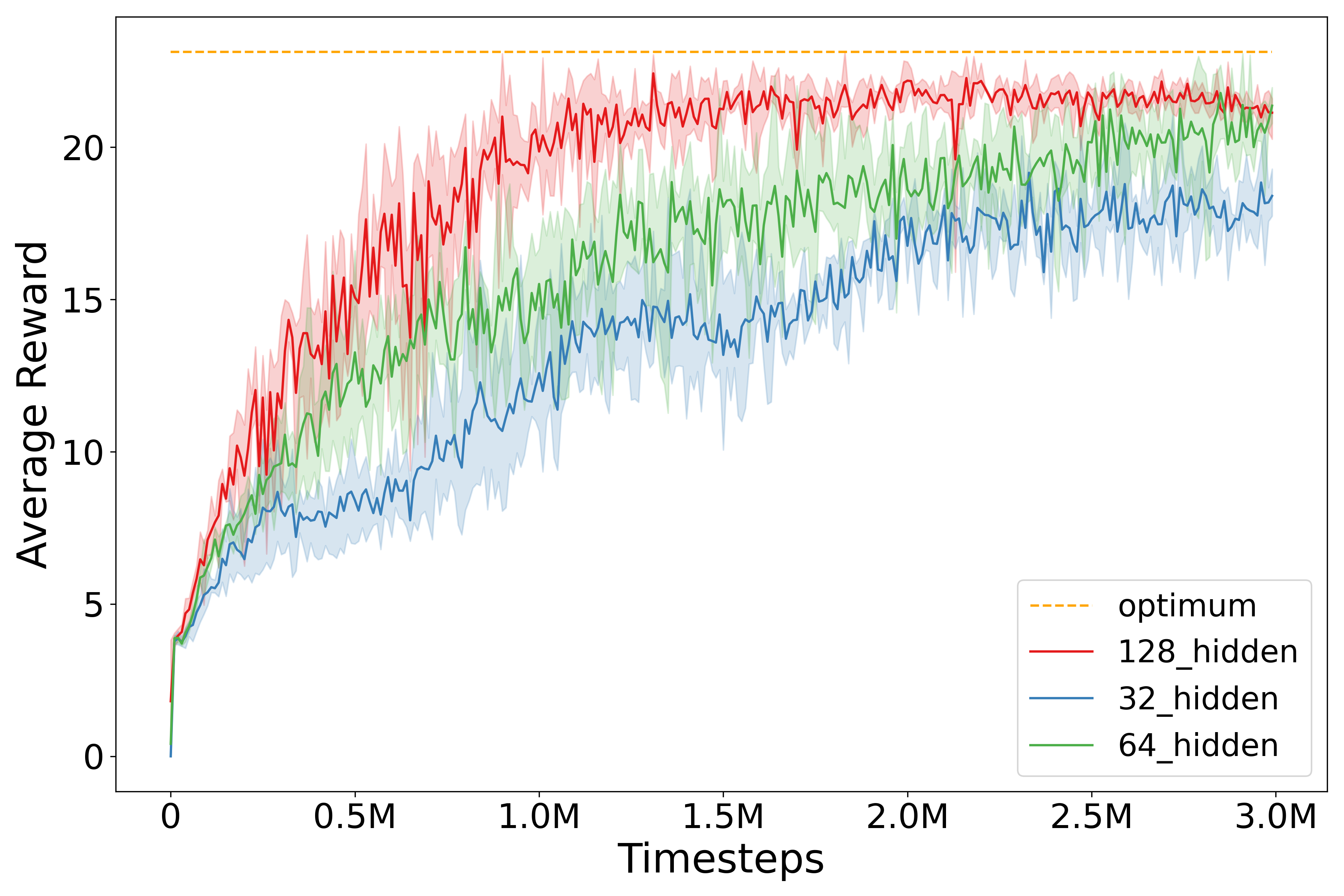}}
    \subfigure[PS+IQL+COMM]{\label{fig:app_r_smac_b}\includegraphics[width=0.23\textwidth]{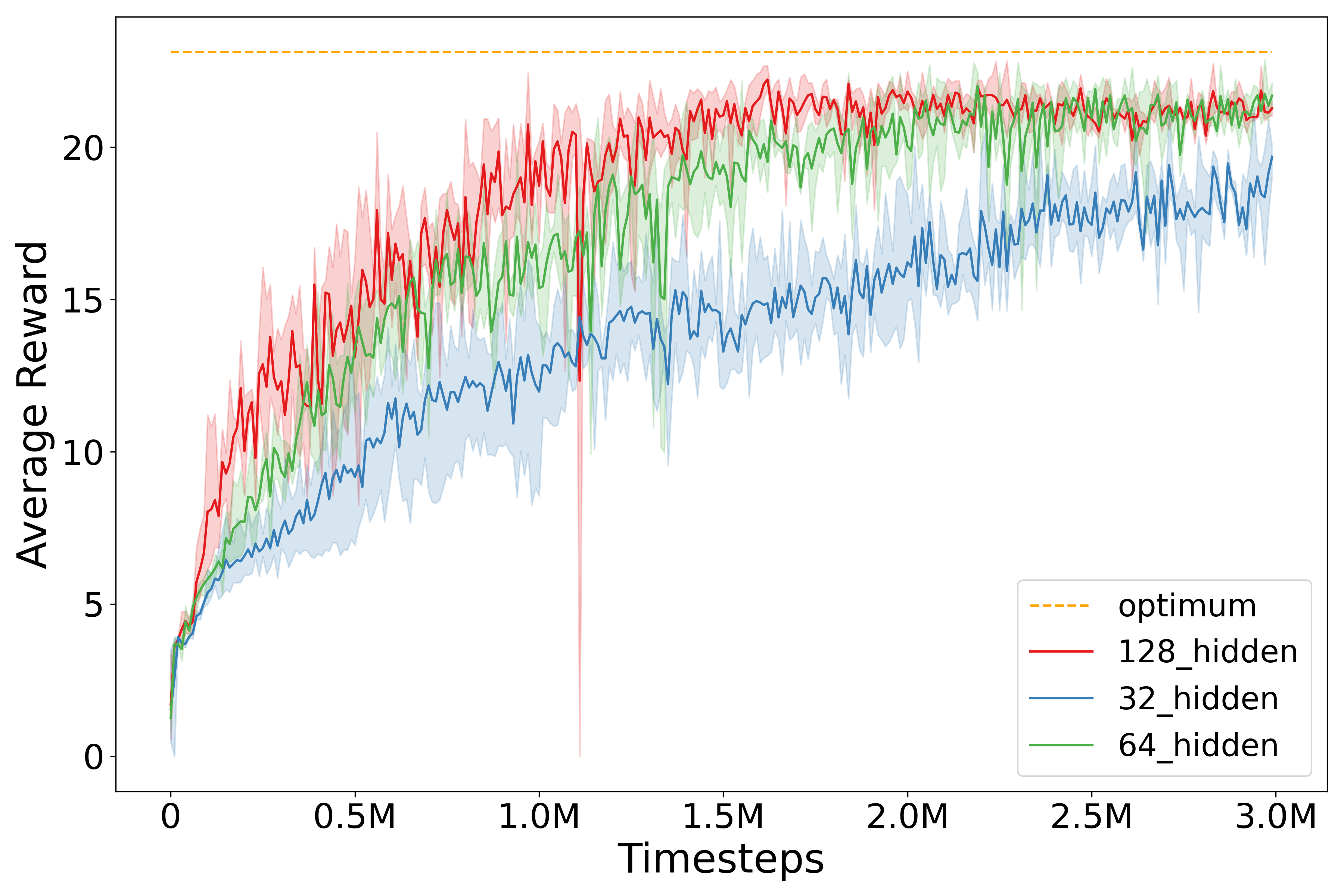}}
    \subfigure[NPS+IQL]{\label{fig:app_r_smac_c}\includegraphics[width=0.23\textwidth]{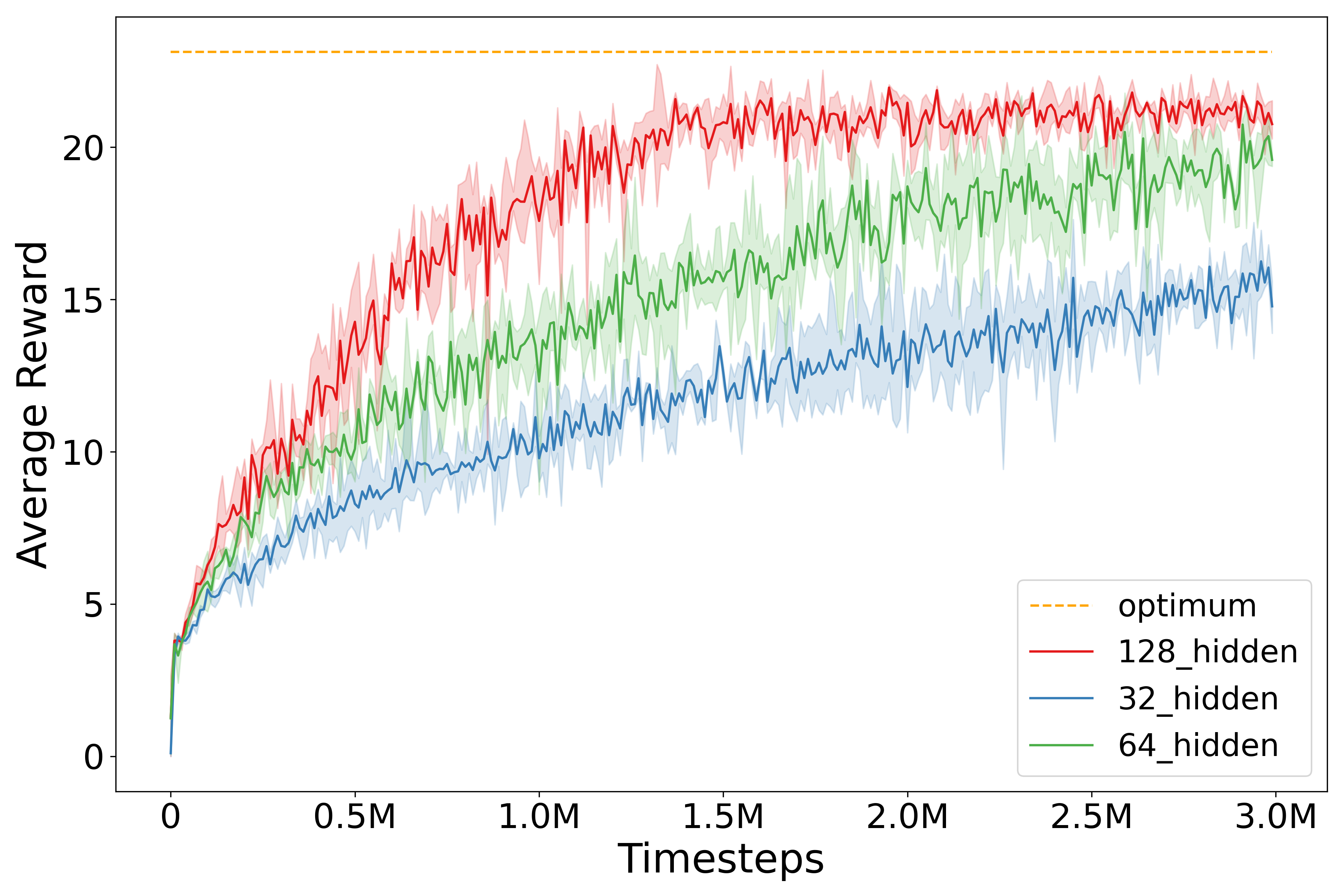}}
    \subfigure[NPS+IQL+COMM]{\label{fig:app_r_smac_d}\includegraphics[width=0.23\textwidth]{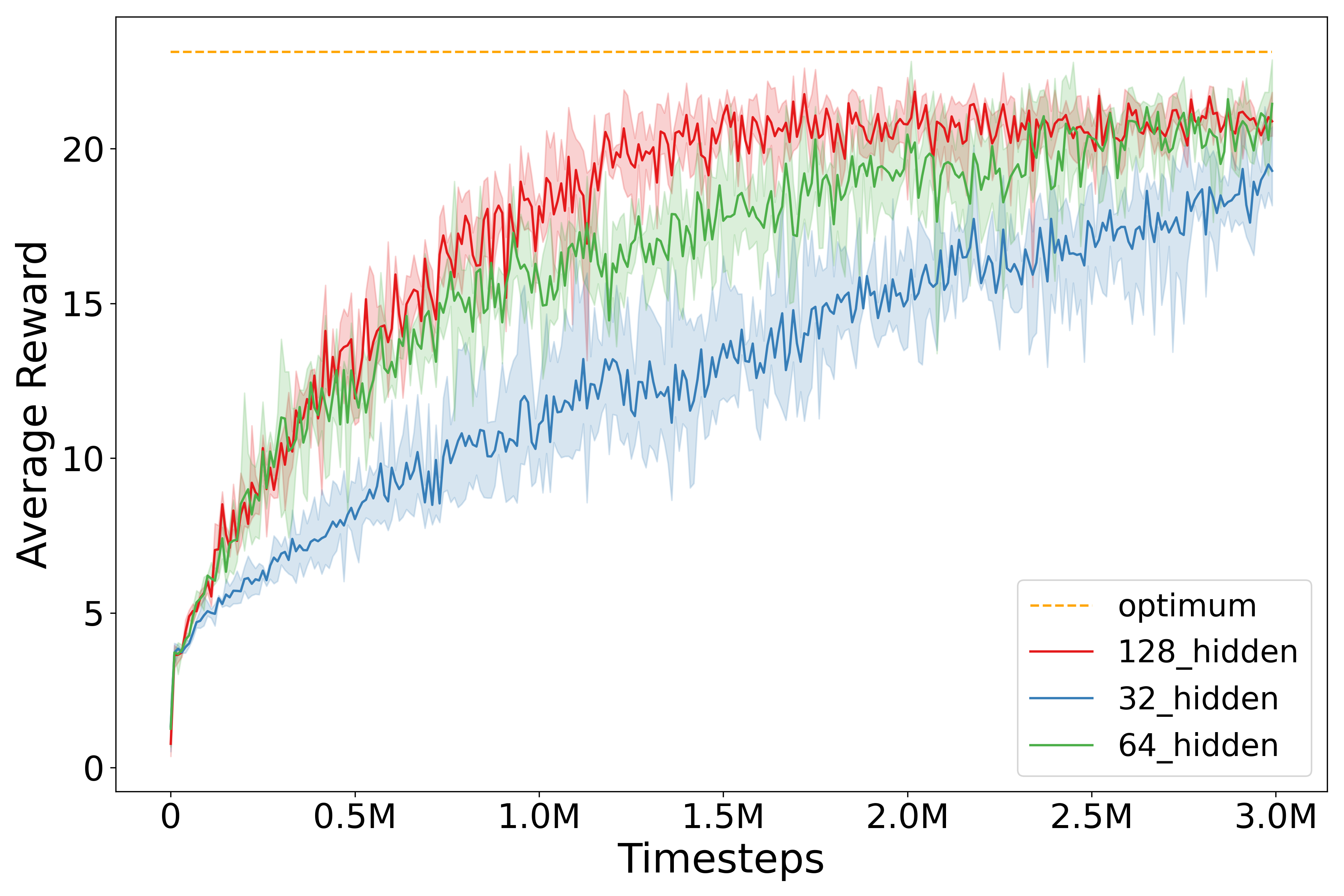}}
    \caption{Rewards achieved by the attempted methods in 3s\_vs\_5z. The dashed line (optimum) represents the maximum value achieved among all the methods in this scenario. It is plotted simply for better visualisation purposes.}
    \label{fig:app_rewards_3svs5z}
\end{figure}

For completeness purposes, in Figure \ref{fig:app_rewards_3svs5z} we also show the resulting rewards for 3s\_vs\_5z that correspond to the win rates in the experiments section of the main paper. We can see that the plots of the rewards tell the same story as the corresponding win rates. However, in some cases, the rewards for smaller network capacities are close to the others with higher capacity, meaning that these agents with lower capacity are likely to be close to better performances. For instance, in Figure \ref{fig:app_r_smac_a} we can see that the performance with size 32 is getting very close to the one with 64. Also in Figure \ref{fig:app_r_smac_d}, we can see that, at the end of training, the agents with size 32 are very close to the ones with size 64 and even the ones with size 128. This suggests that it is likely that they would reach the same optimal performance if they were trained for longer. This leads to the insight that agents with lower capacity will require more training time, but they might reach the same optimal performance if trained for long enough. These observations cannot be easily seen in the plots of the win rates.

\section{Architecture of Independent Learning without Communication}
\begin{figure}[!hbt]
    \centering
    \includegraphics[width=0.6\textwidth]{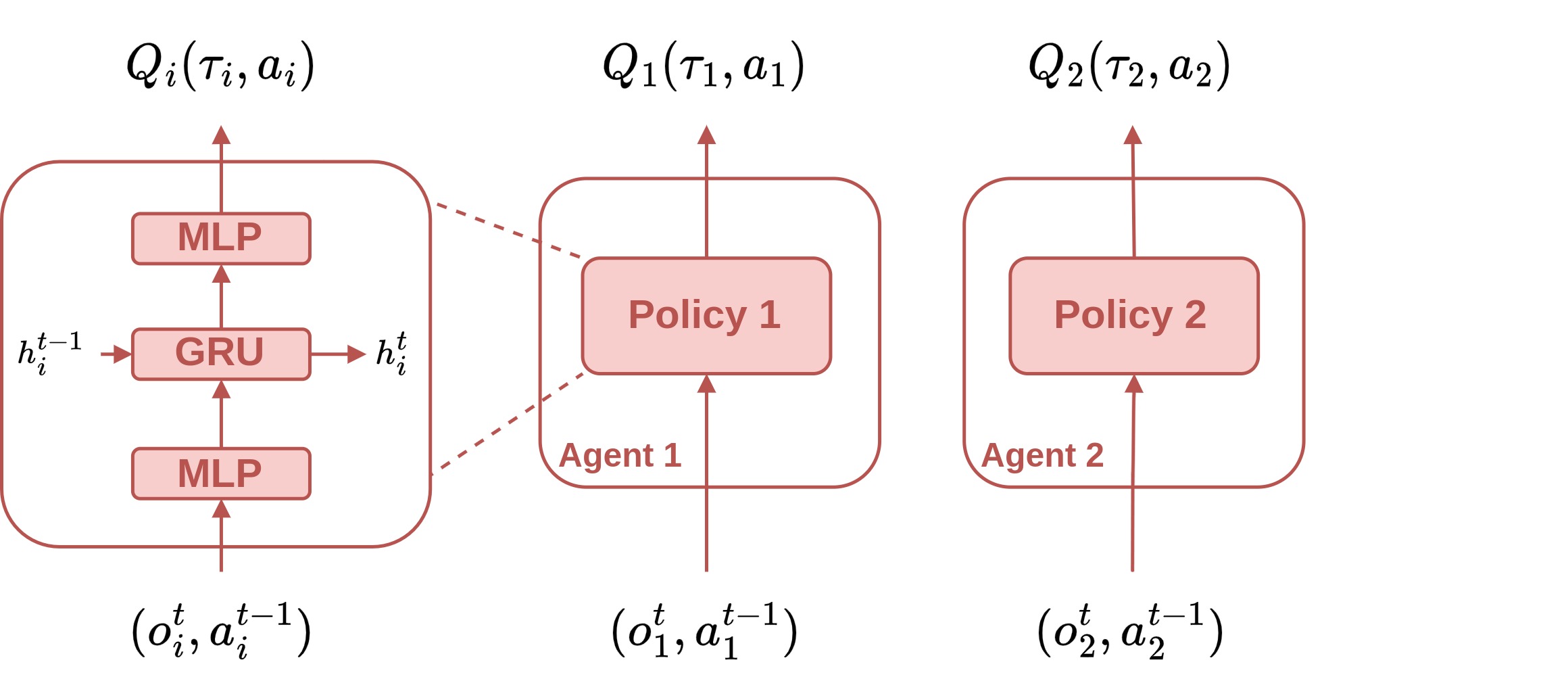}
    \caption{Architecture for NPS+IQL as described in the main text of this paper. When compared to NPS+IQL+COMM, now there is no communication network and the actions of the agents are taken based only on the observations, and not on the messages as well, as shown for NPS+IQL+COMM in section \ref{sec:indep_comm_meth} in the main paper.}
    \label{fig:app_ps_iql_arch}
\end{figure}
For comparison with the architecture in section \ref{sec:indep_comm_meth} of the main paper, in Figure \ref{fig:app_ps_iql_arch} we show the architecture of independent learning without parameter sharing that was used in the experiments in this paper. Intuitively, independent learning with parameter sharing can be deduced from this architecture, if we look at the networks of the agents as if they represent the same network that receives an additional agent ID as input.

\section{Hyperparameters and Implementation Details}
In the experiments presented in the main paper, all the results represent the average of 3 independent runs. In our configuration of deep independent Q-learning, all the agents are controlled by recurrent deep neural networks with a Gated Recurrent Unit (GRU) cell. The default width of the GRU is 64. However, as described in the experiments, we increase or decrease this value to conduct our analysis. We use both parameter sharing and no parameter sharing in the experiments. When parameters are not shared, each individual agent is controlled by a separate network, like the one described. In the experiments with communication, we use a communication neural network based on linear transformations that encodes the observations of the agents. We fix the dimension of the hidden layers to 64. Importantly, when we use parameter sharing, an agent ID is included in the inputs of the agents. However, this is not used anymore when parameters are not shared, since it becomes redundant.

The exploration-exploitation tradeoff of the agents is made according to the epsilon-greedy method, with an initial epsilon of $\epsilon=1$ that anneals down to t a minimum of 0.05 throughout 50000 training episodes. We use the RMSprop optimiser to train all the networks, with a learning rate $\alpha=5\times 10^{-4}$. The discount factor used is $\gamma=0.99$, and the maximum size of the replay buffer is 5000, from which minibatches of 32 episodes are sampled. Every 200 training episodes, the target networks are updated.


\end{document}